\documentclass{article} %
\usepackage{iclr2027_conference,times}

\usepackage{amsmath,amsfonts,bm}

\def\eqref#1{equation~\ref{#1}}

\def\1{\bm{1}}

\DeclareMathAlphabet{\mathsfit}{\encodingdefault}{\sfdefault}{m}{sl}
\SetMathAlphabet{\mathsfit}{bold}{\encodingdefault}{\sfdefault}{bx}{n}

\usepackage{wrapfig}
\usepackage{capt-of}
\usepackage{xcolor}
\definecolor{paperblue}{HTML}{3F75B5}
\usepackage[colorlinks=true,
  citecolor=paperblue,
  linkcolor=paperblue,
  urlcolor=paperblue]{hyperref}
\usepackage[capitalize,nameinlink]{cleveref}
\usepackage{url}
\usepackage{graphicx}
\usepackage{array}
\usepackage{booktabs}
\usepackage{multirow}
\usepackage{xcolor}
\usepackage{algorithm}
\usepackage{algpseudocode}
\usepackage{comment}

\newcommand{\cfigref}[2][]{\hyperref[#2]{Figure~\ref*{#2}#1}}
\newcommand{\ctabref}[2][]{\hyperref[#2]{Table~\ref*{#2}#1}}

\title{HorizonFlow: Variable-Length Planning for Offline Goal-Conditioned RL}

\author{\begin{tabular}{@{}c@{\hspace{1.2em}}c@{\hspace{1.2em}}c@{}}
\href{https://jh2525.github.io/}{JunHyeok Oh}$^{1}$ & \href{https://zianjang.github.io/}{Zian Jang}$^{1}$ & \href{https://scholar.google.com/citations?user=FwoohI4AAAAJ\&hl=en}{Byung-Jun Lee}$^{1,2,\dagger}$\\[-1pt]
{\small\href{mailto:the2ndlaw@korea.ac.kr}{\texttt{the2ndlaw@korea.ac.kr}}} & {\small\href{mailto:jangzian@korea.ac.kr}{\texttt{jangzian@korea.ac.kr}}} & {\small\href{mailto:byungjunlee@korea.ac.kr}{\texttt{byungjunlee@korea.ac.kr}}}
\end{tabular}}

\newcommand{\ms}[2]{#1\,{\tiny$\pm$#2}}

\iclrfinalcopy %
\begin{document}

\vspace*{-0.30in}\maketitle\lhead{}
\footnotetext[1]{Korea University. $^{2}$Gauss Labs Inc., Seoul, Republic of Korea. $^{\dagger}$Corresponding author.}
\vspace{-0.03in}

\begin{abstract}
Recent advances in generative planning have made trajectory inpainting a promising approach to offline goal-conditioned reinforcement learning. However, these methods typically specify the planning horizon before generating plan content, even though the appropriate horizon depends on the route itself. A horizon that is too short can force infeasible transitions, whereas one that is too long can introduce redundant motion. We introduce HorizonFlow, a hierarchical planner that treats plan length as an output of generation rather than a prescribed input. Its subgoal route planner guides its action-prefix controller through a sequence of latent subgoals. Both components combine insertion-based generation with flow matching to jointly generate continuous plan content and length, using the partially generated plan to guide token insertion. HorizonFlow reuses the resulting length information to select candidates and steer generation toward shorter plans without a separate learned value model. Across Maze2D, Multi2D, and OGBench navigation and visual manipulation benchmarks, HorizonFlow achieves the highest average performance among the compared methods.
\end{abstract}
\vspace{-12pt}
{\centering\small\href{https://jh2525.github.io/HorizonFlow/}{\texttt{Project Page: https://jh2525.github.io/HorizonFlow/}}\par}
\vspace{-6pt}
\begin{figure}[h]
  \centering
  \includegraphics[width=\linewidth]{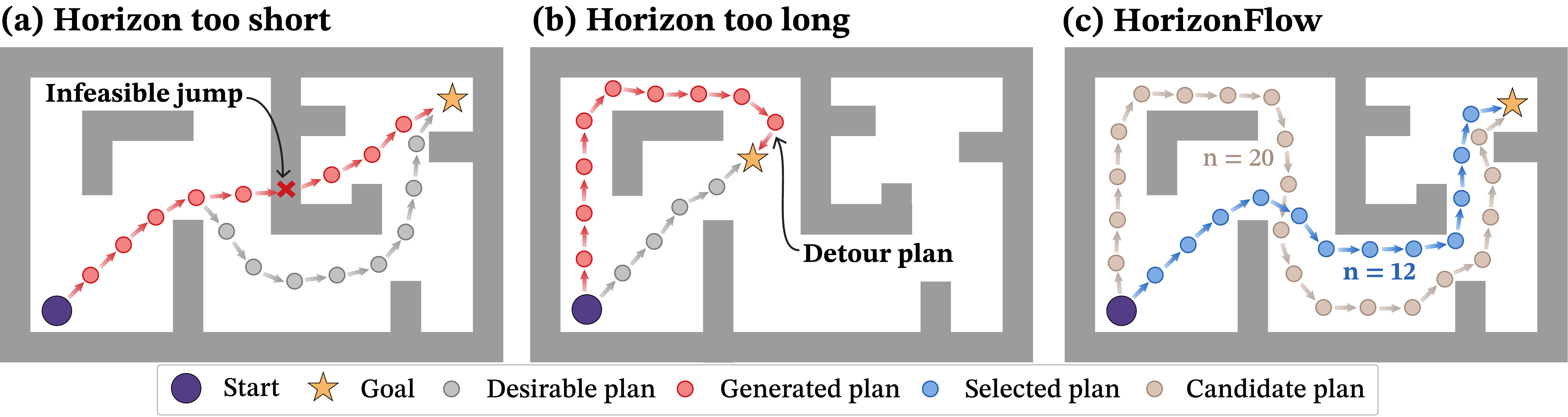}
  \par\vskip-8pt
  \caption{\textbf{Schematic of horizon mismatch and count-based selection.}
(a,~b) Prescribed horizons that are too short or too long yield an
infeasible jump or detour, respectively. (c) HorizonFlow generates
variable-length candidates and selects the one with the fewest tokens.}
  \label{fig:two-blockers}
  \vskip-8pt
\end{figure}

\section{Introduction}
Offline goal-conditioned reinforcement learning (GCRL) aims to reach specified goals using only a fixed dataset, without additional environment interaction. Generative trajectory planning is a natural approach because it models distributions over trajectories from logged experience and can represent alternative ways of reaching a goal as explicit candidates. A common formulation is trajectory inpainting, which fixes the current state and a goal or subgoal as endpoints and generates the trajectory between them~\citep{janner2022diffuser,li2023hdmi,chen2024simple,liu2025vhdiffuser}. In these methods, the planning horizon---the length of the generated plan---is typically specified before generation. Because both endpoints are fixed, this horizon is not a lookahead but an assumed travel time, and it can be wrong in either direction. A horizon that is too short may compress a long trajectory into dynamically infeasible transitions, whereas one that is too long may stretch a short trajectory through waiting, redundant motion, or unnecessary detours. \cfigref[~(a, b)]{fig:two-blockers} illustrates this two-sided horizon mismatch, also noted by \citet{liu2025vhdiffuser}.

Many generative planners support configurable horizons but require them to be selected externally~\citep{janner2022diffuser,chen2024diffusionforcing,luo2025compdiffuser}. Because the appropriate horizon varies across state--goal pairs, no single value fits both nearby and distant goals. VH-Diffuser~\citep{liu2025vhdiffuser} addresses this with a learned length predictor that assigns each start–goal pair a horizon before generation begins. This still requires choosing a useful horizon before the candidate route is available; a poorly matched horizon can constrain the generated plan to be too short or unnecessarily long.

We instead generate plan content and length jointly by combining insertion-based generation with flow matching~\citep{lipman2023flowmatching,nguyen2025oneflow}. Conditioned on the partially generated plan, the insertion model predicts gapwise missing-token counts to determine how many tokens to add, while flow matching refines their continuous content. Length decisions thus depend on each candidate's evolving content rather than on a horizon prescribed before generation. At a fixed temporal resolution, the generated token count reflects the duration a plan represents, which lets us compare candidates without a separate learned value model. During generation, we combine the current count with the predicted missing count into a count-based steering score that favors shorter candidates; after generation, the realized count selects among completed plans.

Building on this joint content–length generation, we introduce HorizonFlow, a hierarchical planner for offline GCRL. Applying the formulation directly to an entire route at action-level resolution would require long sequences, which makes generation challenging. HorizonFlow therefore applies it at two levels. Given the current state and final goal, the subgoal route planner generates a variable-length latent subgoal route. Given the current state and the next subgoal, the action-prefix controller generates a variable-length action prefix. Both models use the same insertion–flow matching formulation and are trained entirely on offline data. HorizonFlow selects candidates by their generated counts at each level and operates in a receding-horizon manner, executing an initial portion of the selected action prefix and replanning as the state evolves. 

Our contributions are threefold. First, we extend insertion-based generation with flow matching to continuous-valued plans, jointly generating plan content and length for control. Second, we introduce HorizonFlow, a hierarchical planner that applies this formulation to subgoal routes and action prefixes, with length-aware sampling that uses generated counts for candidate selection and count-based scores for steering during generation. Third, HorizonFlow achieves the highest average performance among the compared methods on Maze2D, Multi2D, and OGBench navigation and visual manipulation benchmarks; horizon diagnostics and ablations further analyze horizon mismatch, length signals, and the planner's sampling and hierarchy choices.

\section{Related Work}
\label{sec:related}

\paragraph{Offline Goal-Conditioned RL.}
Value-based offline GCRL uses contrastive objectives, quasimetric
representations, or expectile
regression~\citep{eysenbach2022crl,wang2023qrl,park2025ogbench}.
HIQL uses latent-subgoal hierarchies~\citep{park2023hiql}, SAW learns a flat policy through
subgoal-conditioned bootstrapping~\citep{zhou2025saw}, and CTA composes
task-relevant latent analogies with new contexts~\citep{kim2026cta}.
HorizonFlow instead generates variable-length plans and selects by
generated count without learning a value function.

\paragraph{Generative Planning for Control.}
Diffuser and Decision Diffuser generate
trajectories~\citep{janner2022diffuser,ajay2023decisiondiffuser};
hierarchical diffusion/flow planners generate subgoals and local
plans~\citep{li2023hdmi,chen2024simple,nandiraju2026hdflow}.
SSD conditions sub-trajectories on goals and learned values~\citep{kim2024ssd}.
Horizons are specified before content generation through configurable
lookahead in Diffusion Forcing~\citep{chen2024diffusionforcing},
fixed-length chunks in CompDiffuser~\citep{luo2025compdiffuser}, or
prediction in VH-Diffuser~\citep{liu2025vhdiffuser}.
HorizonFlow jointly generates each candidate's content and length.

\paragraph{Variable-Length Generative Models.}
Beyond fixed-length flows~\citep{lipman2023flowmatching,gat2024discreteflow},
jump diffusion and Edit Flows model variable-dimensional and
variable-length outputs, respectively~\citep{campbell2023jumpdiffusion,havasi2025editflows}.
OneFlow combines insertion and flow matching for continuous image
latents within discrete text~\citep{nguyen2025oneflow}; the Insertion
Process plans discrete maze and graph paths over a discrete
vocabulary~\citep{zhang2026insertion}.
HorizonFlow instead jointly generates continuous plan tokens,
including their order coordinates, and token counts, reusing length
information for selection and steering.

\section{Preliminaries}
\label{sec:prelim}

\paragraph{Offline Goal-Conditioned RL.}
\label{sec:prelim-gcrl}

We consider a reward-free Markov decision process
$\mathcal{M}=(\mathcal{S},\mathcal{A},f,\rho_0)$ with state space
$\mathcal{S}$, action space $\mathcal{A}$, deterministic dynamics
$s_{h+1}=f(s_h,a_h)$, and initial-state distribution $\rho_0$.
Extending HorizonFlow to
stochastic dynamics is left for future work. A goal $g\in\mathcal{G}$ specifies successful states through a goal map
$\phi:\mathcal{S}\rightarrow\mathcal{G}$; success occurs when
$\phi(s_h)\in B_\epsilon(g)$, the $\epsilon$-ball around $g$.
In offline GCRL, the learner receives only a static dataset of
trajectories $\mathcal{D}=\{\tau^{(i)}\}_{i=1}^{N_{\mathrm{traj}}}$, where
$\tau=(s_0,a_0,s_1,a_1,\ldots)$, and must reach arbitrary goals without
additional environment interaction~\citep{park2025ogbench}.

\paragraph{Flow Matching.}
\label{sec:prelim-fm}

Flow Matching (FM) learns to transform noise into a continuous data
sample~\citep{lipman2023flowmatching}. In our planner, it generates the
content of subgoal or action tokens, where a token is one element
of a plan. For a noise sample $x_0\sim p_0$, a clean target $x_1\sim q$,
and refinement time $t\sim\mathrm{Unif}[0,1]$, we form the interpolation
$x_t=(1-t)x_0+t x_1$. The model learns the direction $x_1-x_0$ that
moves this noisy sample toward its target, minimizing
$\mathcal{L}_{\mathrm{FM}}=\mathbb{E}_{t,x_0,x_1}
\big[\lVert v_\theta(x_t,t)-(x_1-x_0)\rVert_2^2\big]$.
Samples are obtained by integrating
$\dot{x}_t=v_\theta(x_t,t)$ from $t=0$ to $1$. Standard FM operates on
a fixed-dimensional space. Representing a sequence as
$x_t\in\mathbb{R}^{L\times d}$ therefore fixes its element count $L$;
FM generates element content but does not by itself generate sequence
length.
\paragraph{Insertion-based generation.}
\label{sec:prelim-ef}
An insertion process constructs a variable-length sequence by
repeatedly adding elements between existing neighbors.
Each such location is called a \emph{gap}, and a learned
insertion rate controls how likely an addition is to occur there;
equivalently, the model can predict how many elements each gap
still lacks.
The sequence length is therefore determined by the insertions
made during generation rather than specified in advance.
Edit Flows provides this construction as an insertion-only
restriction of its general sequence-editing
framework~\citep{havasi2025editflows}.
Insertion can be combined with flow matching so that insertion
grows the sequence while flow matching refines the continuous
content of its elements~\citep{nguyen2025oneflow}.

\begin{figure}[t]
  \centering
  \includegraphics[width=\linewidth]{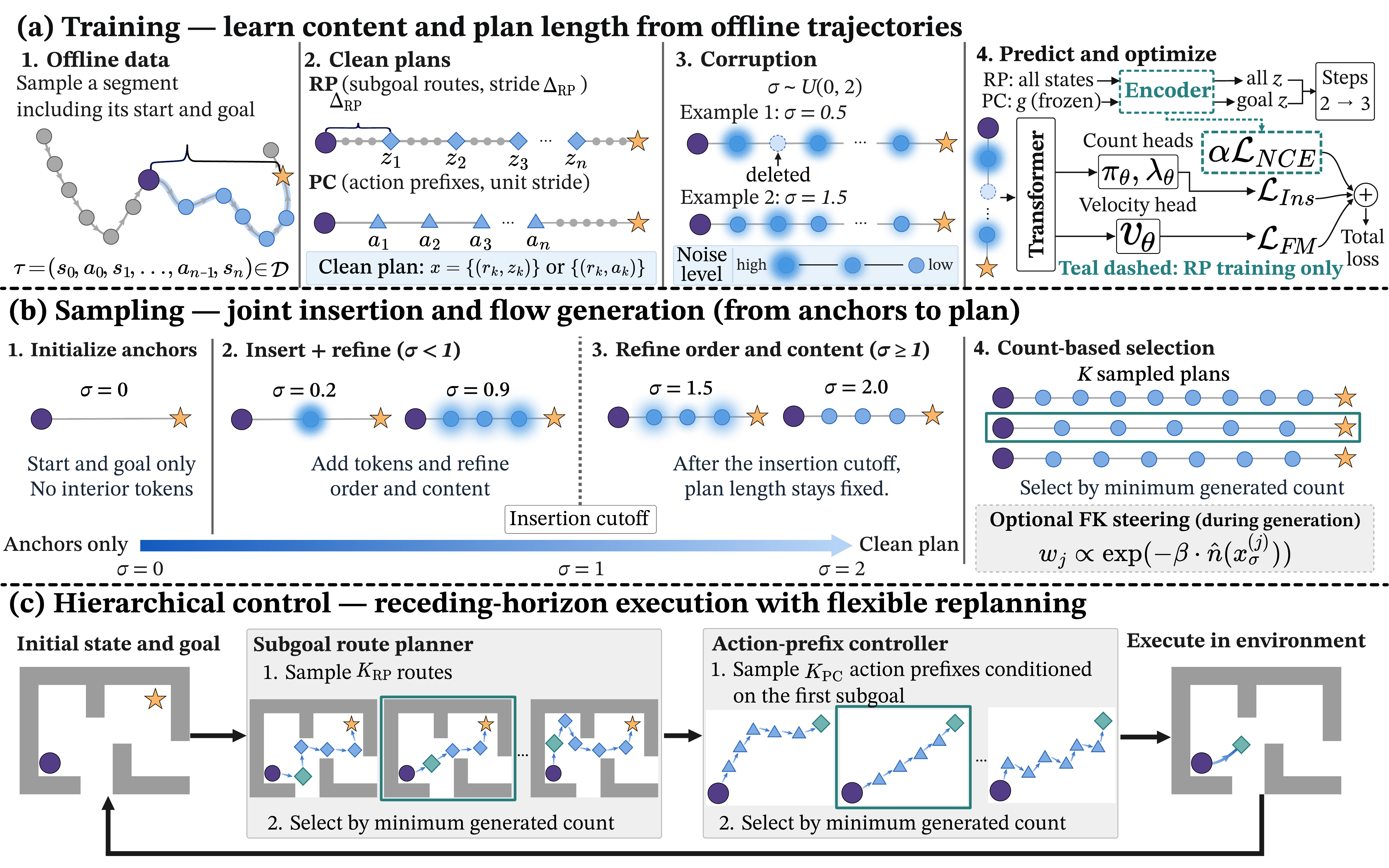}
  \par\vskip-6pt
  \caption{\textbf{HorizonFlow overview.}
(a) Training predicts missing-token counts and denoising velocities
from deleted and noised offline plans.
(b) Sampling inserts and refines tokens before $\sigma=1$, then only
refines; optional Feynman--Kac (FK) steering guides generation,
followed by minimum-count selection.
(c) The route planner (RP) selects a route; the prefix controller (PC)
selects an action prefix toward its first subgoal, with receding-horizon execution.}
  \label{fig:horizonflow-overview}
  \vskip-8pt
\end{figure}
\section{HorizonFlow}
\label{sec:method}

HorizonFlow combines a \emph{subgoal route planner} (RP), which generates
routes to the final goal, with an \emph{action-prefix controller} (PC),
which generates local actions toward the next subgoal.
Both generate variable-length candidates and select by minimum
generated count, with local execution and replanning as the state
changes (\cfigref{fig:horizonflow-overview}).
\Cref{app:method} gives implementation details and pseudocode.

\subsection{Hierarchical Plans and Token Representation}
\label{sec:method-repr}

\paragraph{Plans as token sets.}
Both levels represent a plan as a variable number of continuous
tokens between two fixed conditioning anchors (current state and
goal). Each token $x_k=(r_k,c_k)$ consists of a scalar coordinate
$r_k\in\mathbb{R}$ and content $c_k\in\mathbb{R}^{d}$---a latent
subgoal for the route planner and an action for the prefix
controller. A plan is thus a variable-size set
$x\in\mathcal{X}=\bigcup_{n=0}^{N}(\mathbb{R}^{1+d})^{n}$, with $N$
the token capacity; $|x|$ denotes its token count, which we also
write $n$. The anchors are not counted, remain fixed during
generation, and are excluded from the flow-matching loss.

\paragraph{Order coordinates.}
To order a growing token set, each token carries a continuous order coordinate \(r_k\), which is refined by flow matching together with the
content $c_k$. The anchors sit at $r=-1$ and $r=+1$, and sorting by
$r_k$ orders the interior tokens between them. The coordinate
specifies order along a plan, not elapsed environment time or
flow-matching time. Token content and temporal resolution depend on
the planning level, as described below.

\paragraph{Subgoal route planner.}
Given the current state and final goal, the route planner generates
an ordered sequence of latent subgoals---states mapped by a learned
encoder $E_\phi:\mathcal{S}\to\mathbb{R}^{d_z}$
(\Cref{sec:method-train}) to intermediate points along a
route. Its anchors carry the current and goal states encoded by the same encoder, and its
generated interior tokens are $x=\{(r_k,z_k)\}_{k=1}^{n}$ with latent
subgoals $z_k\in\mathbb{R}^{d_z}$.
Its training targets are states sampled at a fixed stride
$\Delta_{\mathrm{RP}}$ in environment
steps~\citep{chen2024simple}.
The stride sets the temporal resolution, not the number of subgoals:
different routes can contain different numbers of intermediate points.
The first subgoal of the selected route becomes the prefix
controller's target.

\paragraph{Action-prefix controller.}
Conditioned on the current state and the selected latent subgoal,
carried by its start and goal anchors, the prefix controller
generates ordered action tokens $x=\{(r_k,a_k)\}_{k=1}^{n}$ with
$a_k\in\mathbb{R}^{d_a}$, where $n$ is the generated
action-prefix length. After sorting by $r_k$, the action tokens
form a sequence at unit environment-step resolution.
This sequence is a local action prefix toward the subgoal,
not necessarily a complete trajectory to it.
The route planner determines where to go; the prefix controller
models how to begin moving toward that target.
Unlike a one-step inverse dynamics model conditioned on adjacent
states, it generates an action sequence conditioned on a potentially
more distant subgoal.

\subsection{Training}
\label{sec:method-train}

From corrupted offline-trajectory segments, HorizonFlow learns to
predict missing-token counts and denoise surviving tokens.
Encoder regularization and local-prefix training support hierarchical
control.

\paragraph{Learning routes from offline trajectories.}
We first sample a clean token count $m$, including anchors,
log-uniformly between a minimum count and $N+2$; this
emphasizes shorter plans while retaining longer examples. We then
take a segment of $(m-1)\Delta_{\mathrm{RP}}$ environment steps
within one episode, use its endpoints as the (start, goal) anchors,
and subsample its interior every $\Delta_{\mathrm{RP}}$ steps to
obtain the clean RP token set $x_1$.

\paragraph{Corruption and generation times.}
We use $\sigma\in[0,2]$ for the global clock that advances plan
generation. Each token has its own FM refinement time
$t_k\in[0,1]$. For training corruption, each non-anchor token $k$
is assigned an independent reveal offset $u_k\sim\mathrm{Unif}[0,1]$.
We set $t_k=\mathrm{clip}_{[0,1]}(\sigma-u_k)$ and hide tokens for
which $\sigma<u_k$. Thus, for $\sigma<1$ some clean tokens may be
hidden; for $\sigma\ge1$ all are present and only refinement remains.
Neither clock is the order coordinate $r_k$.
The vector $\mathbf{t}$ collects these local times.
We write $x_\sigma$ (equivalently $x_{\mathbf t}$) for the partially
noised plan at clock $\sigma$, both in the training losses and for a
candidate at a steering checkpoint.
\Cref{app:corruption} gives the full corruption procedure.

\paragraph{Learning insertion and continuous content.}
At sampled $\sigma$, hidden clean tokens define each gap's missing
count $c_g$, while surviving tokens are noised to their local times.
A shared Transformer predicts continuous velocities $v_\theta$,
positive-count parameters $\lambda_\theta$, and gap-completion
probabilities $\pi_\theta$ from the noisy plan $x_\sigma$.
We suppress this conditioning in $\pi_\theta(g)$ and $\lambda_\theta(g)$
unless needed. For a present non-anchor token $k$ with clean target
$x_{1,k}=(r_k,c_k)$ and noise $\epsilon_k$, the FM loss is schematically
\begin{equation}
  \mathcal{L}_{\mathrm{FM}}
  = \mathbb{E}\!\left[
    \left\lVert
      v_\theta(x_{\mathbf{t}},\mathbf{t})_k-(x_{1,k}-\epsilon_k)
    \right\rVert_2^2
  \right].
  \label{eq:fm-loss-main}
\end{equation}
Time weighting, coordinate normalization, and loss reduction are
detailed in \Cref{app:training-details}.
The insertion heads model gap completion ($c_g=0$) with a Bernoulli
probability $\pi_\theta(g)$ and positive missing counts with a
zero-truncated Poisson (ZTP) parameter $\lambda_\theta(g)$, giving
\begin{equation}
  \mathcal{L}_{\mathrm{ins}}
  = \mathbb{E}_{g}\!\left[
    \mathrm{BCE}\!\left(\pi_\theta(g),\mathbf{1}[c_g=0]\right)
    + \mathbf{1}[c_g>0]\bigl(
      \lambda_\theta(g)-c_g\log\lambda_\theta(g)
      +\log(1-e^{-\lambda_\theta(g)})
    \bigr)
  \right].
  \label{eq:ins-loss-main}
\end{equation}
The normalizer $\log(1-e^{-\lambda})$ makes the positive-count branch
a proper distribution over $c_g\geq1$; the parameter-independent
term $\log c_g!$ is omitted. This normalization matters because,
besides supervising insertions, the same distribution provides the
expected remaining count used by the optional steering step in
\Cref{sec:method-fk}. \Cref{app:oneflow} relates the
objective to OneFlow's zero/nonzero count decomposition.

\paragraph{Stabilizing the subgoal encoder.}
The encoder $E_\phi$ is trained jointly with the route planner at
this stage and frozen thereafter. Its outputs must retain
distinctions between states to serve as useful subgoals, but the FM
loss alone rewards collapsing them: a less distinct latent makes the
velocity target easier to fit. We therefore block FM gradients at
the encoder and train it through the insertion loss and temporal
contrastive regularization. The InfoNCE term encourages nearby-in-time
states to have similar representations relative to in-batch
negatives~\citep{oord2018cpc}; it supports the representation rather
than providing a separate ranking score. With $\alpha=0.1$ in all
environments, the RP objective is
\begin{equation}
  \mathcal{L}_{\mathrm{RP}} \;=\; \mathcal{L}_{\mathrm{FM}}
    + \mathcal{L}_{\mathrm{ins}}
    + \alpha\, \mathcal{L}_{\mathrm{NCE}}.
  \label{eq:total-loss}
\end{equation}
\paragraph{Keeping local generation short.}
The prefix controller is trained with
$\mathcal{L}_{\mathrm{PC}}=\mathcal{L}_{\mathrm{FM}}+\mathcal{L}_{\mathrm{ins}}$
on action tokens, using the frozen encoder $E_\phi$ for goal
conditioning and no contrastive loss. A subgoal can lie farther
away than the controller's token capacity, especially at unit
stride in long-horizon environments, so the controller should not
have to generate every action up to it. We therefore pair a
training prefix of $n=m-2$ actions, with $m$ sampled log-uniformly, with
a hindsight goal $K_{\mathrm{div}}\,n$ steps ahead: with
$K_{\mathrm{div}}=1$ the prefix reaches the goal frame, whereas with
$K_{\mathrm{div}}>1$ it covers only the first $1/K_{\mathrm{div}}$
of the way. The controller thus learns to make progress toward a
distant subgoal without modeling the entire remaining route in
action space. \Cref{app:training-details} gives, for both
planners, the exact prefix construction, token budgets, corruption
schedule, masking, and loss weighting.

\subsection{Sampling Variable-Length Plans}
\label{sec:method-sample}

\begin{figure}[t]
  \centering
  \includegraphics[width=\linewidth]{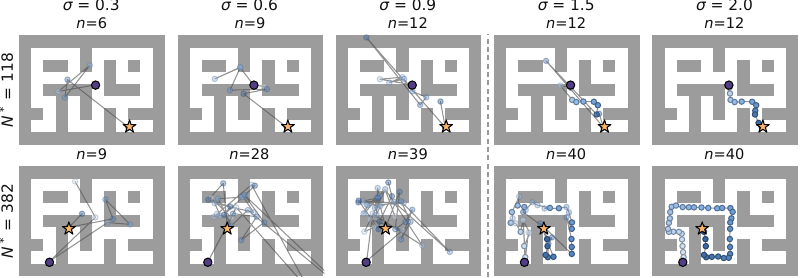}
  \par\vskip-6pt
  \caption{\textbf{Route-planner samples on Maze2D-Large.}
Rows show start--goal pairs; columns show generation time $\sigma$
and generated count $n$. Latent subgoals are mapped to maze coordinates,
connected and colored by order coordinate $r_k$.
White circles/orange stars mark starts/goals.
Beyond the dashed cutoff at $\sigma=1$, counts remain fixed while
refinement continues. These are generated plans, not executed trajectories.}
  \label{fig:xy-plan-build}
  \vskip-8pt
\end{figure}

\cfigref[(b)]{fig:horizonflow-overview} illustrates sampling in two
phases. Each candidate starts with fixed start and goal anchors and no
interior tokens. During the insertion phase ($\sigma<1$), the model
predicts a flow velocity for every active token and the gapwise
completion probability $\pi_\theta$ and count parameter
$\lambda_\theta$ for every gap. Gapwise birth decisions drawn from
these determine how many new tokens are added; each is initialized
from Gaussian noise at local time $t_k=0$, while flow matching
refines the order coordinates and content of existing tokens.

Births stop at $\sigma=1$, fixing $n$; refinement continues to
$\sigma=2$, giving even a token born just before the cutoff a full
refinement interval.
The same sampler generates RP subgoals and PC actions, with counts
used for selection (\Cref{sec:method-plan}).
\Cref{app:method:sample} gives the sampling rule and pseudocode;
\cfigref{fig:xy-plan-build} illustrates generation on Maze2D-Large. %

\paragraph{Optional Feynman--Kac steering.}
\label{sec:method-fk}
We reuse the insertion heads to steer generation toward shorter plans
through Feynman--Kac resampling~\citep{singhal2025fk}. For a partial
candidate $x_\sigma$, the count-based steering score
(\Cref{app:count-derivation}) is
\begin{equation}
  \hat{n}(x_\sigma) = \underbrace{|x_\sigma|}_{\text{tokens present}} +
  \underbrace{\textstyle\sum_{g \in G(x_\sigma)}
    \bigl(1 - \pi_\theta(g \mid x_\sigma)\bigr)
    \frac{\lambda_\theta(g \mid x_\sigma)}
         {1-e^{-\lambda_\theta(g \mid x_\sigma)}}}%
    _{\text{predicted missing tokens}},
  \label{eq:nhat}
\end{equation}
where $G(x_\sigma)$ is the set of gaps. At a few checkpoints during
the insertion phase, we resample candidates with weights
$w_j\propto\exp[-\beta\hat n(x_\sigma^{(j)})]$, favoring lower
scores without an auxiliary trajectory-value model.
FK can be applied at either level; by default we apply it to the
route planner and the prefix controller. Implementation details are in
\Cref{app:method:sample}.

\begin{table}[t]
\caption{\textbf{Benchmark performance.}
Values are mean $\pm$ standard deviation; highlighted entries mark
row-wise highest means. HorizonFlow uses 5, 8, and 4 training seeds for
Maze2D/Multi2D, OGBench navigation, and visual manipulation, respectively,
with 100 episodes per environment and protocol for Maze2D/Multi2D
and 50 per task for OGBench.}
\label{tab:benchmarks}
\centering
\vskip4pt
\begin{minipage}{\linewidth}
\centering
\textbf{(a) Maze2D / Multi2D: D4RL normalized score}\par
\scriptsize
\setlength{\tabcolsep}{2.2pt}
\renewcommand{\arraystretch}{1.0}
\resizebox{\linewidth}{!}{%
\begin{tabular}{@{}ll*{7}{c}@{}}
\toprule
\textbf{Environment} & \textbf{Task} & Diffuser & VHD & HDMI & HD & DF & SSD & \textbf{Ours} \\
\midrule
\multirow{3}{*}{\texttt{Maze2D}}
 & U-Maze & \ms{113.9}{3.1} & \ms{118.5}{6.7} & \ms{120.1}{5.6} & \ms{128.4}{36.0} & \ms{116.7}{2.0} & \textbf{\color{paperblue}\ms{144.6}{7.6}} & \ms{137.5}{0.8} \\
 & Medium & \ms{121.5}{2.7} & \ms{130.5}{3.6} & \ms{121.8}{3.6} & \ms{135.6}{30.0} & \ms{149.4}{7.5} & \ms{134.4}{13.6} & \textbf{\color{paperblue}\ms{154.8}{1.5}} \\
 & Large  & \ms{123.0}{6.4} & \ms{142.9}{7.1} & \ms{128.6}{6.5} & \ms{155.8}{25.0} & \ms{159.0}{2.7} & \ms{183.5}{19.2} & \textbf{\color{paperblue}\ms{205.2}{2.4}} \\
\midrule
\multicolumn{2}{l}{\textbf{Single-task Average}} & 119.5 & 130.6 & 123.5 & 139.9 & 141.7 & 154.2 & \textbf{\color{paperblue}165.8} \\
\midrule
\multirow{3}{*}{\texttt{Multi2D}}
 & U-Maze & \ms{128.9}{1.8} & \ms{137.6}{3.9} & \ms{131.3}{4.0} & \ms{144.1}{12.0} & \ms{119.1}{4.0} & \textbf{\color{paperblue}\ms{158.2}{10.1}} & \ms{145.1}{1.2} \\
 & Medium & \ms{127.2}{3.4} & \ms{146.3}{2.0} & \ms{131.6}{4.2} & \ms{140.2}{16.0} & \ms{152.3}{9.9} & \ms{155.2}{17.7} & \textbf{\color{paperblue}\ms{170.5}{1.0}} \\
 & Large  & \ms{132.1}{5.8} & \ms{169.4}{3.6} & \ms{135.4}{5.6} & \ms{165.5}{6.0} & \ms{167.1}{2.7} & \ms{192.9}{19.0} & \textbf{\color{paperblue}\ms{216.5}{1.0}} \\
\midrule
\multicolumn{2}{l}{\textbf{Multi-task Average}} & 129.4 & 151.1 & 132.8 & 149.9 & 146.2 & 168.8 & \textbf{\color{paperblue}177.4} \\
\bottomrule
\end{tabular}
}
\end{minipage}
\par\medskip
\begin{minipage}{\linewidth}
\centering
\textbf{(b) OGBench navigation: success rate (\%)}\par
\footnotesize
\setlength{\tabcolsep}{2.0pt}
\renewcommand{\arraystretch}{1.0}
\begin{tabular*}{\linewidth}{@{\extracolsep{\fill}}ll*{9}{c}@{}}
\toprule
\textbf{Environment} & \textbf{Task}
& \multicolumn{5}{c}{\textbf{Value-based}}
& \multicolumn{4}{c}{\textbf{Planning-based}} \\
\cmidrule(lr){3-7}\cmidrule(l){8-11}
& & QRL & CRL & HIQL & SAW & CTA & Diffuser & HD & DF & \textbf{Ours} \\
\midrule
\multirow{3}{*}{\texttt{Pointmaze}}
 & Medium & \ms{82}{5} & \ms{29}{7} & \ms{79}{5} & \ms{97}{2} & \ms{87}{4} & \ms{95}{3} & \ms{95}{2} & \ms{82}{8} & \textbf{\color{paperblue}\ms{100.0}{0.0}} \\
 & Large  & \ms{86}{9} & \ms{39}{7} & \ms{58}{5} & \ms{85}{10} & \ms{71}{12} & \ms{99}{0} & \ms{98}{1} & \ms{60}{3} & \textbf{\color{paperblue}\ms{99.8}{0.2}} \\
 & Giant  & \ms{68}{7} & \ms{27}{10} & \ms{46}{9} & \ms{68}{8} & \ms{30}{14} & \ms{92}{4} & \ms{96}{3} & \ms{52}{12} & \textbf{\color{paperblue}\ms{97.1}{3.0}} \\
\midrule
\multirow{3}{*}{\texttt{Antmaze}}
 & Medium & \ms{88}{3} & \ms{95}{1} & \ms{96}{1} & \ms{97}{1} & \ms{96}{1} & \ms{77}{2} & \ms{46}{7} & \ms{32}{4} & \textbf{\color{paperblue}\ms{97.3}{1.2}} \\
 & Large  & \ms{75}{6} & \ms{83}{4} & \ms{91}{2} & \ms{90}{3} & \ms{85}{3} & \ms{61}{3} & \ms{19}{8} & \ms{4}{3} & \textbf{\color{paperblue}\ms{93.3}{1.1}} \\
 & Giant  & \ms{14}{3} & \ms{16}{3} & \ms{65}{5} & \ms{73}{4} & \ms{54}{4} & \ms{5}{3} & \ms{1}{1} & \ms{0}{0} & \textbf{\color{paperblue}\ms{86.5}{2.9}} \\
\midrule
\multirow{3}{*}{\texttt{Humanoidmaze}}
 & Medium & \ms{21}{8} & \ms{60}{4} & \ms{89}{2} & \ms{88}{3} & \ms{90}{2} & \ms{39}{3} & \ms{67}{2} & \ms{25}{3} & \textbf{\color{paperblue}\ms{98.6}{0.4}} \\
 & Large  & \ms{5}{1} & \ms{24}{4} & \ms{49}{4} & \ms{46}{4} & \ms{60}{3} & \ms{6}{2} & \ms{18}{2} & \ms{3}{2} & \textbf{\color{paperblue}\ms{80.2}{1.5}} \\
 & Giant  & \ms{1}{0} & \ms{3}{2} & \ms{12}{4} & \ms{35}{4} & \ms{5}{1} & \ms{1}{0} & \ms{7}{3} & \ms{0}{0} & \textbf{\color{paperblue}\ms{95.5}{1.3}} \\
\midrule
\multicolumn{2}{l}{\textbf{Average}} & 48.9 & 41.8 & 65.0 & 75.4 & 64.3 & 52.6 & 49.8 & 28.6 & \textbf{\color{paperblue}94.2} \\
\bottomrule
\end{tabular*}%
\end{minipage}
\par\medskip
\begin{minipage}{\linewidth}
\centering
\textbf{(c) OGBench visual manipulation: success rate (\%)}\par
\footnotesize
\setlength{\tabcolsep}{3pt}
\renewcommand{\arraystretch}{1.0}
\begin{tabular*}{\linewidth}{@{\extracolsep{\fill}}l*{8}{c}@{}}
\toprule
\textbf{Environment} & GCIVL & GCIQL & QRL & CRL & HIQL & SAW & CTA & \textbf{Ours} \\
\midrule
\texttt{visual-cube-single} & \ms{60}{5} & \ms{30}{5} & \ms{41}{15} & \ms{31}{15} & \ms{89}{0} & \ms{88}{3} & \ms{89}{2} & \textbf{\color{paperblue}\ms{93.2}{1.1}} \\
\texttt{visual-cube-double} & \ms{10}{2} & \ms{1}{1} & \ms{5}{0} & \ms{2}{1} & \ms{39}{2} & \ms{40}{3} & \ms{8}{2} & \textbf{\color{paperblue}\ms{46.8}{11.3}} \\
\texttt{visual-cube-triple} & \ms{14}{2} & \ms{15}{1} & \ms{16}{1} & \ms{17}{2} & \textbf{\color{paperblue}\ms{21}{0}} & \ms{20}{1} & \ms{4}{3} & \ms{14.4}{2.4} \\
\texttt{visual-scene} & \ms{25}{3} & \ms{12}{2} & \ms{10}{1} & \ms{11}{2} & \ms{49}{4} & \ms{47}{6} & \ms{59}{11} & \textbf{\color{paperblue}\ms{65.6}{0.6}} \\
\midrule
\textbf{Average} & 27.3 & 14.5 & 18.0 & 15.3 & 49.5 & 48.8 & 40.1 & \textbf{\color{paperblue}55.0} \\
\bottomrule
\end{tabular*}%
\end{minipage}
\par
\vskip-6pt
\end{table}

\subsection{Selecting and Executing Plans}
\label{sec:method-plan}
\paragraph{Count-based selection.}
The route planner samples $K_{\mathrm{RP}}$ routes conditioned on
the current state and final goal; the first subgoal of the selected
route then conditions the prefix controller, which samples
$K_{\mathrm{PC}}$ action prefixes from the current state. At either
level, we select among candidates $x^{(1)},\dots,x^{(K)}$ by
\begin{equation}
  x^\star \;=\; x^{(j^\star)},\qquad
  j^\star \;=\; \arg\min_{j \le K} \; |x^{(j)}| ,
  \label{eq:minn}
\end{equation}
counting latent subgoal tokens for RP and action tokens for PC, with
ties broken uniformly at random. At a fixed temporal resolution, token count is a proxy for the temporal extent represented by a plan. We use it to favor compact plans under the learned distribution without a separate trajectory-value model, not to certify feasibility or optimality. For PC, training pairs $n$ actions with a goal $K_{\mathrm{div}}n$ steps ahead, so target count is proportional to the goal offset at fixed $K_{\mathrm{div}}$. We use this learned relation to rank action prefixes before executing the first action and replanning. Unlike prescribing a short horizon, it compares lengths after
the candidates are generated, allowing each route to determine its
own length.

\paragraph{Receding-horizon execution.}
\label{sec:method-exec}
The first subgoal of the selected route serves as the prefix
controller's target, and the route is refreshed every
$H_{\mathrm{RP}}$ environment steps, the RP update period.
Conditioned on this target, the controller executes an initial
portion of its selected action prefix before replanning from the
updated state. By default, the route update period is approximately half the RP training stride, and only the first action is executed per replan. The route update period and the executed prefix fraction
control how often global routes and local actions are revised; we
study both in \cfigref[(c,~d)]{fig:compute-alloc}. Execution details
are given in \Cref{app:method:control}.
\begin{figure}[t]
  \centering
  \includegraphics[width=\linewidth]{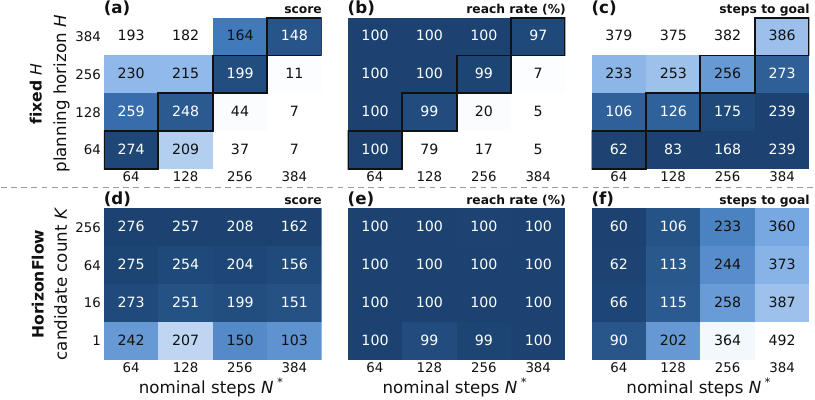}
  \par\vskip-6pt
\caption{\textbf{Planning-horizon and candidate-count diagnostics on Maze2D-Large.}
Both models generate XY routes tracked by a shared
proportional--derivative (PD) controller.
(a--c) Fixed-length (no-insertion) model with prescribed horizon $H$;
outlines mark $H=N^*$.
(d--f) Joint content--length model with FK steering and minimum-count
selection over $K$ candidates.
Panels show normalized score, reach rate, and median steps to goal
among successful rollouts; columns are nominal-step bins $N^*$~(100 pairs each). Darker is better; colors are normalized per column
across both models.}
  \label{fig:horizon-overview}
  \vskip-8pt
\end{figure}
\section{Experiments}
\label{sec:exp}

We evaluate HorizonFlow as a complete planning system on Maze2D,
Multi2D, and OGBench, then examine horizon sensitivity and the effects
of candidate counts, steering, replanning, inference cost, and the
RP--PC hierarchy.
\paragraph{Benchmark comparisons.}
\ctabref[(a--c)]{tab:benchmarks} compares HorizonFlow with prior
methods on Maze2D/Multi2D, OGBench navigation, and visual manipulation.
Maze2D uses a fixed goal (single-task), whereas Multi2D resamples it
each episode (multi-task).
For Maze2D/Multi2D, baselines include
Diffuser~\citep{janner2022diffuser},
HDMI~\citep{li2023hdmi},
Hierarchical Diffuser (HD)~\citep{chen2024simple},
Diffusion Forcing (DF)~\citep{chen2024diffusionforcing},
SSD~\citep{kim2024ssd}, and VH-Diffuser (VHD)~\citep{liu2025vhdiffuser}.
For OGBench, we compare with the value-based methods
QRL~\citep{wang2023qrl}, CRL~\citep{eysenbach2022crl},
HIQL~\citep{park2023hiql}, SAW~\citep{zhou2025saw}, and
CTA~\citep{kim2026cta}, adding GCIVL and GCIQL~\citep{park2025ogbench}
for visual manipulation, and with the planning-based methods Diffuser,
HD, and DF for navigation.
HorizonFlow uses one training and inference recipe whose stride,
update period, and controller capacity scale with each environment's
episode limit (\Cref{app:method}). HorizonFlow attains the highest average in all three groups, with the
largest margins in long-horizon environments.

\paragraph{Sensitivity to the planning horizon.}
We compare the joint content--length model with a separately trained
fixed-length (no-insertion) model, both generating routes in XY space
with a shared Transformer backbone, offline data, training budget, and
proportional--derivative (PD) waypoint controller.
We group start--goal pairs by nominal steps
$N^*(s,g)=d_{\mathrm{geo}}(s,g)/v$, where $d_{\mathrm{geo}}$
is obstacle-avoiding geodesic distance and $v$ is a nominal speed
based on dataset motion (\Cref{app:horizon-diag:nstar}).
Fixed horizons trade off reachability and execution efficiency
(\cfigref[(a--c)]{fig:horizon-overview}): short horizons struggle
with distant goals, whereas long horizons delay arrival at nearby
goals. Joint generation maintains high reach rates across nominal-step bins even with one candidate, without
prescribing a horizon for each start--goal pair
(\cfigref[(d--f)]{fig:horizon-overview}).
Although a single candidate arrives later than the fixed horizon
matched to the privileged $N^*$, selecting among $K \geq 64$
candidates exceeds that horizon's score in every bin.

\paragraph{Candidate counts.}
On OGBench navigation, increasing $K_{\mathrm{RP}}$ raises average
success from 56.2\% to 94.3\% at the default $K_{\mathrm{RP}}=16$,
after which gains level off (\cfigref[(a)]{fig:compute-alloc}).
Since $K=1$ is a single unranked draw, equivalent in distribution to
uniform selection among independent candidates
(\Cref{app:method:sample}), the unsteered gain over $K=1$ is
attributable to minimum-count selection.
FK steering is most useful at small $K_{\mathrm{RP}}$, matching
unsteered success with two to four times fewer candidates.
Increasing $K_{\mathrm{PC}}$ yields smaller gains (92.3\% to 94.6\%;
\cfigref[(b)]{fig:compute-alloc}).

\begin{figure}[t]
  \centering
  \includegraphics[width=\linewidth]{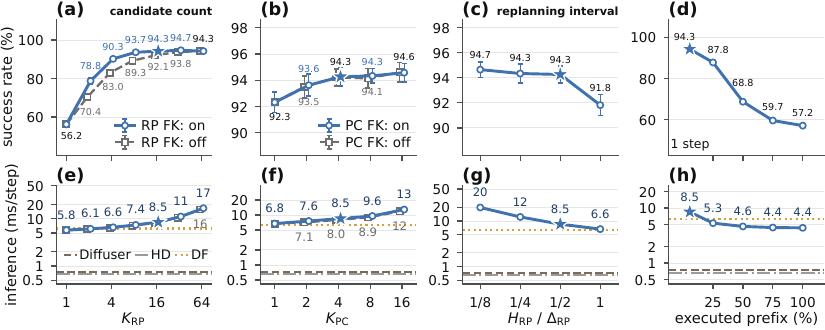}
  \par\vskip-6pt
  \caption{\textbf{Sampling, replanning, and inference cost on OGBench navigation.}
Columns vary RP candidate count (a,~e), PC candidate count (b,~f),
RP update period relative to training stride (c,~g), and executed prefix fraction (d,~h),
with one-step execution shown separately; other settings stay at their defaults (stars).
Rows report mean success (a--d) and estimated amortized model-inference
time (e--h; ms/step, log scale), averaged equally over
the nine navigation environments on a matched evaluation subset.
Solid/dashed curves denote FK steering on/off; horizontal lines in
(e--h) mark baseline timing references.
Error bars combine per-environment standard deviations across seeds
(details in \Cref{app:compute-alloc-domains}).}
  \label{fig:compute-alloc}
  \vskip-8pt
\end{figure}

\paragraph{Selection risk.}
Minimum-count selection can favor a short but infeasible candidate.
In the XY diagnostic, this rarely happens (\Cref{app:horizon-diag}).
The risk is not negligible everywhere: excessively large candidate
counts reduce success in some environments (e.g., visual-cube-triple;
\Cref{app:compute-alloc-domains}), possibly because larger pools are
more likely to contain a short infeasible outlier, consistent with
prior observations in planning~\citep{ki2025priorguided}.
Still, this risk differs in kind from that of an insufficient
prescribed horizon. A horizon must be matched to each start--goal pair
before generation, and no single value suits both nearby and distant
goals (\cfigref[(a--c)]{fig:horizon-overview}). The candidate count,
in contrast, is a single global setting: average success varies by about one
point for $K_{\mathrm{RP}}$ between 8 and 64.

\paragraph{Replanning frequency.}
Between route updates, the controller holds the first subgoal of the
selected route as its target (\Cref{app:method:control}).
On OGBench navigation, success is nearly unchanged for RP update
periods from one-eighth to half of the training stride but drops at
the full stride (94.3\% to 91.8\%;
\cfigref[(c)]{fig:compute-alloc}), possibly because the subgoal is
reached before the next update.
Conversely, longer periods slightly help in some visual manipulation
tasks, possibly because frequent updates shift the target under short strides
(\Cref{app:compute-alloc-domains}).
Executing more of each action prefix before replanning reduces
success more sharply, from 94.3\% to 57.2\%
(\cfigref[(d)]{fig:compute-alloc}), mainly in HumanoidMaze and
AntMaze-Giant, where open-loop errors likely compound quickly.

\paragraph{Inference cost.}
\cfigref[(e--h)]{fig:compute-alloc} estimates amortized model-inference
time per environment step on OGBench navigation.
More candidates increase cost; longer RP update periods reduce it.
At the RP level, FK adds modest fixed-$K$ overhead but can match
unsteered success with fewer candidates and lower estimated cost
(\cfigref[(a, e)]{fig:compute-alloc}).
Diffuser and HD's lower per-step costs partly reflect predominantly
one-shot timing, whereas HorizonFlow's default controller replans
local actions every step. At the default settings, HorizonFlow averages
about 8.5\,ms of model computation per environment step, making frequent
replanning practical in terms of average model-compute cost on the
evaluated hardware.
HorizonFlow has higher controller-only cost but lower planning-update
latency estimates (\Cref{app:step-latency}).
Measurement details and training times are in
\Cref{app:inference-hardware,app:training-time}.

\paragraph{Architecture ablations.}
Replacing the prefix controller with a behavior-cloning policy reduces
average navigation success from 94.2\% to 67.8\%, and removing the
hierarchy reduces it to 42.9\% (\Cref{app:architecture-ablation}).
\section{Conclusion}
We presented HorizonFlow, a hierarchical planner that treats plan length as an output of generation rather than a prescribed input. Insertion-based generation and flow matching jointly generate continuous content and length for both latent subgoal routes and action prefixes. Generated counts support candidate selection without a separate learned value model, while count-based scores steer generation toward shorter plans. HorizonFlow achieves the highest average performance among the compared methods on Maze2D, Multi2D, and OGBench navigation and visual manipulation. Diagnostics show that joint generation reaches both nearby and distant goals without a per-pair horizon and that generated counts closely track goal distance, while ablations confirm the role of both hierarchy levels. Our evaluation focuses on deterministic environments; the effectiveness of count-based selection and steering under stochastic dynamics remains to be established.

\bibliography{iclr2027_conference}
\bibliographystyle{iclr2027_conference}

\appendix
\crefalias{section}{appendix}
\crefalias{subsection}{appendix}
\crefalias{subsubsection}{appendix}

\section{Horizon Diagnostic: Protocol}
\label{app:horizon-diag}

\begin{figure}[t]
  \centering
  \includegraphics[width=\linewidth]{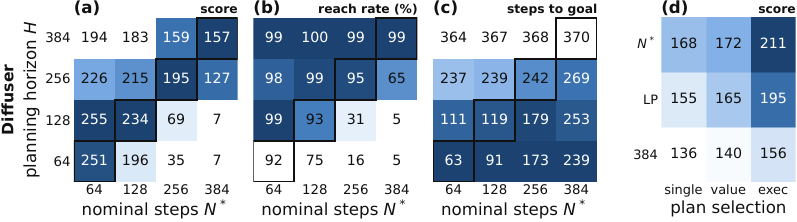}
  \caption{Diffuser diagnostics on Maze2D-Large. (a--c) Normalized score,
  reach rate, and median steps to goal among successful rollouts across
  planning horizon $H$ and nominal steps $N^*$. Outlines mark $H=N^*$.
  (d) Normalized score under different horizon-assignment and selection
  protocols: LP denotes learned length prediction; single, value, and exec
  denote single-candidate, value-based, and execution-based selection.
  Darker is better; colors are normalized within columns in (a--c) and
  across all cells in (d).}
  \label{fig:diffuser-horizon-diagnostic}
\end{figure}

We report three complementary diagnostics on Maze2D-Large.
\cfigref{fig:horizon-overview} compares prescribed-length and joint
content--length XY generation under shared execution conditions.
\cfigref{fig:diffuser-horizon-diagnostic} reports a Diffuser horizon
sweep and a separate horizon-assignment and candidate-selection study.
\cfigref{fig:length-signal-correlation} compares learned length signals.
The models, evaluation populations, and sampling settings are
specified separately below. These diagnostics complement the
full-system benchmark evaluation by examining specific aspects
of horizon assignment and count-based planning.

\subsection{Nominal Steps}
\label{app:horizon-diag:nstar}

For a start--goal pair, we define nominal steps as
\begin{equation}
  N^* = \frac{d_{\mathrm{geo}}(s,g)}{v},
  \label{eq:nominal-travel-time}
\end{equation}
where $d_{\mathrm{geo}}$ is an obstacle-avoiding shortest-path distance
on a discretized free-space map and $v$ is a nominal speed constant
chosen with reference to the dataset. This is a distance scale, not
a measured minimum number of environment steps. A controller can
arrive in fewer steps by moving faster than $v$ or by entering the
goal radius before reaching the goal point.
For Maze2D-Large, positions are in cell units; cell $(i,j)$ is the
unit square centered at $(i,j)$. We use the 4-connected breadth-first
distance between start and goal cells on the $9\times12$ wall map and
$v=0.0385$ cells per step. The measured dataset displacement is 0.0345
cells per step on average, with median 0.0356 and 90th percentile 0.048.

\subsection{XY Horizon and Candidate-Count Diagnostic}
\label{app:horizon-diag:xy}

\paragraph{Diagnostic design.}
This diagnostic examines the consequences of prescribing plan length
before generation versus generating length jointly with content.
We generate routes directly in XY space and execute both variants
with the same PD waypoint controller, using the same backbone
architecture, training budget, and evaluation protocol.
This design removes learned-representation and learned-controller
differences from the comparison, allowing us to examine how the
two generation strategies affect executed route quality under
common execution conditions.

In HorizonFlow's full hierarchy, the state encoder is optimized
jointly with the route planner. Replacing the generation objective
and retraining the encoder can therefore change both the generator
and the representation in which it operates. Freezing a shared
encoder would instead evaluate generation under a fixed
representation, rather than the original joint-learning procedure;
using an encoder learned by either variant could also favor that
variant. Direct XY generation avoids this dependence by providing
a common, explicit representation that is not learned by either model.

We likewise omit the learned prefix controller from this diagnostic.
Its role in the full system is to translate intermediate subgoals
into local action prefixes for feedback control, whereas the question
here concerns route generation and horizon assignment. Including
separately learned controllers would introduce differences in how
generated routes are executed. A common PD controller removes this
source of variation while retaining execution-based evaluation.
The benchmark experiments evaluate the complete planning system,
and the architectural ablations assess its hierarchical and controller
design choices; this diagnostic instead provides a focused comparison
of prescribed-length and joint content--length generation.

\paragraph{Models and training.}
For \cfigref{fig:horizon-overview}, we train two route-generation models
directly on XY coordinates from the Maze2D-Large offline dataset.
Each token contains an order coordinate and a two-dimensional position;
no learned encoder, contrastive objective, or prefix controller is used.
The models share the Transformer backbone, data normalization, training
stride $\Delta_{\mathrm{RP}}=13$, batch size 1,024, and 100,000-update
training budget.
Training samples variable anchor-inclusive counts using the same
log-uniform sampler and storage buckets $\{8,16,32,64\}$.
The insertion model uses the insertion and flow-matching objectives.
The separately trained no-insertion model uses all-present flow-matching
corruption without insertion heads: all tokens for the sampled count are
present from the start. Thus fixed length refers to the count
prescribed at inference, not training at a single length.
The fixed-length condition therefore uses a separately trained
no-insertion generator, rather than disabling insertion only at
inference in the trained insertion model.

\paragraph{Prescribed-length generation.}
\label{app:horizon-diag:xy-fixed}
For \cfigref[(a--c)]{fig:horizon-overview}, we initialize the no-insertion model with an
anchor-inclusive count
\begin{equation}
  m=\operatorname{round}(H/\Delta_{\mathrm{RP}})+1,
\end{equation}
and refine the non-anchor tokens without births, keeping the start and
goal anchors fixed. Each generated route is linearly interpolated at
stride $\Delta_{\mathrm{RP}}$ to form the waypoint reference.
Its time span is $\Delta_{\mathrm{RP}}(m-1)$, so the displayed requests
$H\in\{64,128,256,384\}$ yield spans of 65, 130, 260, and 390 environment
steps, respectively. We draw one candidate per pair and horizon.
Both models use 20 global generation steps.

\paragraph{Adaptive generation and selection.}
\label{app:horizon-diag:xy-adaptive}
For \cfigref[(d--f)]{fig:horizon-overview}, the insertion model generates the count jointly with
the route content. The $K=1$ condition uses the first candidate from an
unsteered pool. Each displayed multi-candidate condition
$K\in\{16,64,256\}$ uses its own FK-steered pool, with resampling at
generation steps 3 and 7, strength $\beta=0.5$, and the hurdle--ZTP
count-based steering score. We then select the minimum generated
non-anchor count. These rows evaluate FK steering together with final
count-based selection, not an isolated selection-only intervention.

\paragraph{Execution and aggregation.}
The two models use the same start--goal pairs, nominal-step bins,
800-step execution cap, and PD waypoint controller defined in
\Cref{app:horizon-diag:grid}. Each reference route is generated once;
there is no hierarchical replanning. The figure displays 100 pairs in
each of four bins from the shared six-bin set.
Ties in minimum count receive equal weight within each pair.
Scores and reach rates average these pairwise expectations over the
100 pairs; steps to goal is the weighted median among successful
tied candidates, using the same weights and excluding failures.
The single-candidate rows reduce to ordinary means and
successful-rollout medians. Each variant is a single trained model.

\paragraph{Display and scope.}
For each metric and nominal-step column, colors are min--max normalized
jointly over the displayed fixed-length and adaptive rows; steps to goal
is negated so that darker means fewer steps. This normalization also
applies to reach rate; printed percentages remain on their original scale.
Outlines indicate equality between the requested $H$ and bin center,
before stride rounding.
The comparison removes encoder and learned-controller differences,
but the two generators have different corruption processes and objectives.
It therefore examines prescribed versus jointly generated length in
this XY setting. The $N^*$ condition supplies privileged geometric length information.

\paragraph{Within-pair count ranking.}
\label{app:horizon-diag:within-pair}
To distinguish ranking candidates for the same problem from ranking
different start--goal pairs by difficulty, we execute the candidates
within each generated pool and compare their counts with their outcomes.
\Cref{tab:within-pair-count} reports 600 pools per configuration.
For each pool, minimum-count selection chooses uniformly among all
candidates tied for the smallest generated count. Its expected success
is therefore the fraction of successful candidates in this tied set,
averaged across pools, rather than the outcome of one sampled tie-break.
The missed-success rate averages the corresponding failure probability
over pools containing at least one successful candidate.
These metrics retain failed candidates in the selection evaluation.

\begin{table}[t]
\centering
\caption{Within-pair execution diagnostics for count-based selection.
Success is the expectation under uniform minimum-count tie-breaking;
available counts pools containing a successful candidate, and miss
reports expected selection failure conditional on that availability.
Median $\rho$ summarizes within-pool Spearman correlations between
count and steps to goal among successful candidates.
$\Delta$ steps compares minimum-count selection with the mean steps
to goal of successful candidates in the same pool; negative values
denote earlier arrival.}
\label{tab:within-pair-count}
\begin{tabular*}{\linewidth}{@{\extracolsep{\fill}}lrrrrr@{}}
\toprule
Pool & \shortstack{Success\\(\%)} & Available &
\shortstack{Miss\\(\%)} & \shortstack{Median\\$\rho$} &
\shortstack{$\Delta$\\steps} \\
\midrule
FK-off, $K=16$  & 99.86  & 600/600 & 0.14 & 0.82 & $-104$ \\
FK-on, $K=16$   & 100.00 & 600/600 & 0.00 & 0.87 & $-42$ \\
FK-on, $K=64$   & 99.94  & 600/600 & 0.06 & 0.84 & $-54$ \\
FK-on, $K=128$  & 100.00 & 600/600 & 0.00 & 0.83 & $-58$ \\
FK-on, $K=256$  & 99.92  & 600/600 & 0.08 & 0.82 & $-63$ \\
\bottomrule
\end{tabular*}
\end{table}

\paragraph{Selection reliability and execution efficiency.}
Every evaluated pool contains a successful candidate, and minimum-count
selection retains high expected success while rarely choosing a failing
candidate over a successful alternative.
Among successful candidates for the same pair, larger counts are
associated with more steps to goal, and the reported step differences
favor minimum-count selection over the successful-candidate mean.
The unsteered condition exhibits the same positive association,
so this within-pair evidence is not restricted to FK-steered pools.
Unlike the across-pair comparison in
\Cref{app:horizon-diag:estimators}, this diagnostic directly relates
count to alternative executed plans for a fixed start and goal.

The correlation and step comparison are conditional on successful
execution; the success and missed-success columns separately expose
selection failures rather than dropping them from the evaluation.
These results support count as a useful ranking signal in this
diagnostic, not a guarantee that the shortest generated candidate is
feasible or optimal.

\subsection{Diffuser Horizon Grid on Maze2D-Large}
\label{app:horizon-diag:grid}

\cfigref[(a--c)]{fig:diffuser-horizon-diagnostic} reports the
Diffuser horizon sweep. Its start--goal pairs and execution controller
are also used for the XY diagnostic in \Cref{app:horizon-diag:xy}.

\paragraph{Start--goal pairs.}
We form six bins centered at $c\in\{64,128,192,256,320,384\}$ nominal
steps, each with half-width 32 steps. A cell pair belongs to bin $c$
when
\begin{equation}
  (c-32)v \leq d_{\mathrm{geo}}(s,g) < (c+32)v.
\end{equation}
The bins contain 246, 464, 366, 470, 222, and 158 candidate pairs,
respectively. From each bin we sample 100 pairs uniformly without
replacement using a fixed random-generator seed, giving 600 pairs.
Start and goal states use cell-center positions and zero velocity.
The column labels report bin centers, not the exact $N^*$ of every
pair in a column.

\paragraph{Planner and horizon settings.}
We use Diffuser~\citep{janner2022diffuser} with its public Maze2D-Large
weights, trained with horizon 384 and sampled with 256 diffusion
steps. We vary the prescribed planning horizon
$H\in\{64,128,192,256,320,384\}$ by changing the sequence length,
without retraining the fully convolutional U-Net. In this diagnostic,
$H$ counts predicted state slots, including the endpoints, rather
than the number of transitions between those slots. Sampling uses
DDPM with $x_0$ clipped to $[-1,1]$ and the start and goal states
inpainted at the first and last positions after every denoising step.
We draw one plan per pair and horizon, yielding
$6\times6\times100=3{,}600$ plans. The execution procedure uses the
predicted positions and finite-difference reference velocities.

\paragraph{Execution and score (a).}
We use $h$ to index environment steps.
Each plan is tracked for 800 environment steps in
\texttt{maze2d-large-v1} without replanning. The waypoint sequence is
fixed, but the tracking controller uses feedback from the observed
position and velocity. We use the Diffuser waypoint controller with
the PD gains used in Diffusion Forcing~\citep{chen2024diffusionforcing}:
\begin{equation}
  a_h = \operatorname{clip}_{[-1,1]}\!\left(
    12.5\,(w_{h+1}-p_h)
    +1.2\,(\dot w_{h+1}-\dot p_h)
  \right),
\end{equation}
where $w_{h+1}$ is the next waypoint, $\dot w$ its finite-difference
velocity, and $p_h,\dot p_h$ the observed position and velocity.
After exhausting the plan, the controller holds the last waypoint
with zero target velocity. Within the goal radius of 0.5, it switches
to the goal-holding rule $a_h=10(g-p_h)-\dot p_h$.
The sparse reward gives one point per environment step inside this
radius. Panel (a) reports the D4RL-normalized return
\begin{equation}
  100\,\frac{R-6.7}{273.99-6.7},
\end{equation}
averaged over the 100 pairs in each horizon--distance cell.

\paragraph{Reach rate (b).}
For each plan, we record the first rollout index at which the agent
is within distance 0.5 of the goal after an environment step. A plan
is counted as reached if this occurs during the 800-step rollout.
Panel (b) reports the percentage of the 100 plans in each cell that
reach the goal. This is an execution-based measure; no waypoint or
line-segment wall test filters these rollouts.

\paragraph{Steps to goal (c).}
Panel (c) reports the median number of executed environment actions
until first arrival among successful rollouts in each cell. Failed
rollouts are excluded; every displayed cell contains at least one
successful rollout.

\paragraph{Interpreting the decomposition.}
The panels distinguish failure to reach from delayed arrival among
successful rollouts; they do not measure wall-crossing frequency.
Steps to goal can fall below a column's $N^*$ because each bin spans
a range of distances, the controller can exceed the nominal pace,
and reaching uses a goal radius. The figure displays four of the
six measured horizon settings and distance bins without pooling
the omitted bins. Its outlines mark equality with a bin center,
not an oracle-optimal horizon for each pair.

\subsection{Diffuser Horizon Assignment and Candidate Selection}
\label{app:horizon-diag:selection}

\cfigref[(d)]{fig:diffuser-horizon-diagnostic} is a separate experiment
using five reproduced Diffuser training seeds, with the LP row using the
corresponding VHD models. These models are not the public checkpoint
used in panels (a--c).
Rows compare a fixed-horizon baseline with $H=384$, a learned
length-prediction variant based on VHD (LP), and a variant using nominal
steps $N^*$. The LP condition follows VHD's length-prediction mechanism
and associated training changes; it should not be interpreted as simply
attaching a predictor to an otherwise identical Diffuser model. Columns use
a single sample, value-based selection using an XY HIQL model among four
candidates, or an execution-based selection oracle among 128 candidates. The execution-based
selection oracle chooses the earliest successful arrival after executing
all candidates from the same initial state, falling back to the first
candidate if none reaches the goal. It is an execution-based reference,
not an offline selection rule. For the $N^*$--execution cell, the oracle additionally searches
the horizon ladder $\{0.75,0.875,1,1.125,1.25\}N^*$.
Each condition uses five training seeds and 100 evaluation episodes
per seed for each of the single-goal and multi-goal settings.
The two settings are averaged within each seed and then across seeds;
scores are D4RL-normalized. Panel (d) uses a single color scale across
its cells and displays means without error bars.

Panel (d) follows the VHD evaluation protocol and is intended as a
within-panel comparison of horizon-assignment and selection strategies,
not as a continuation of the horizon grid in panels (a--c). Its nominal-step estimate uses an
8-connected geodesic calculation on a 0.2-unit grid with nominal speed
0.025 units per environment step. For the ordinary $N^*$ condition,
the resulting horizon is rounded up to a multiple of 32 and clipped
to $[32,384]$. The $N^*$--execution ladder applies its multipliers to
this clipped base, rounds each candidate horizon to the nearest
multiple of 4, and clips it to $[4,384]$. Multiple ladder settings
can therefore coincide at the cap.
Its waypoint tracker uses unit gains on position and velocity errors,
instead of the gains 12.5 and 1.2 used in panels (a--c).

For the LP condition, the single-sample and value-based variants replan after exhausting a
plan without reaching the goal. On replanning, the new horizon is
at least twice the previous plan length, capped by the maximum
horizon and rounded to a multiple of 32. Its execution-based variant
disables replanning so that each candidate is evaluated as one
complete reference plan.

\begin{figure}[t]
  \centering
  \includegraphics[width=\linewidth]{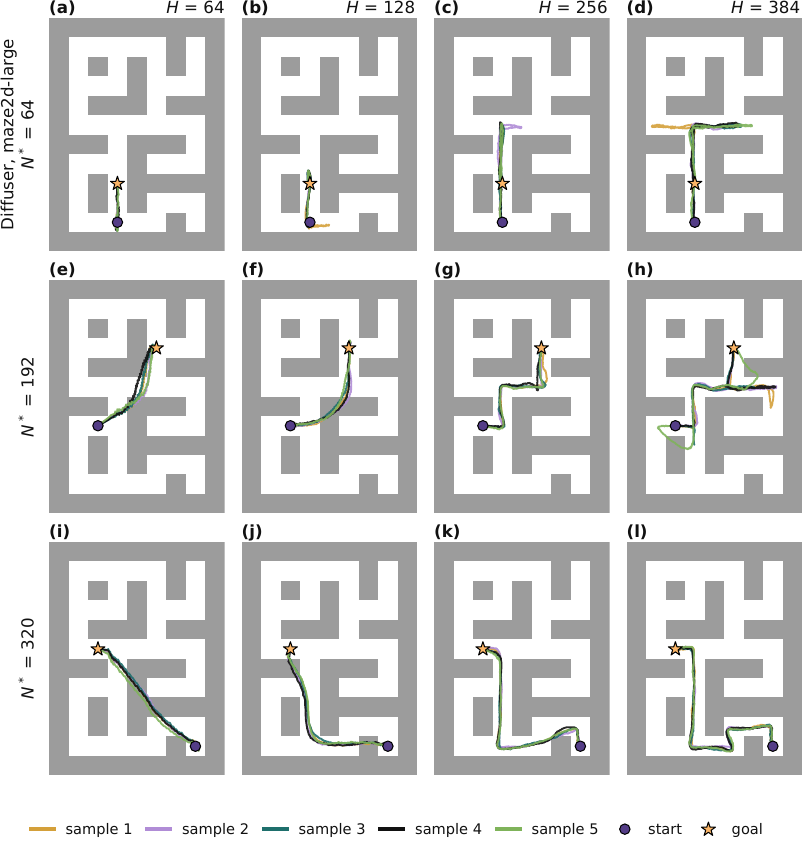}
  \caption{Diffuser plan samples on Maze2D-Large. Rows show
  start--goal pairs labeled by nominal steps
  $N^*\in\{64,192,320\}$; columns vary the planning horizon
  $H\in\{64,128,256,384\}$. Each panel overlays five generated
  position sequences with the same endpoints. Circles mark starts,
  stars mark goals, and gray regions are walls. These are generated
  reference plans, not executed trajectories.}
  \label{fig:diffuser-plan-samples}
\end{figure}

\begin{figure}[t]
  \centering
  \includegraphics[width=\linewidth]{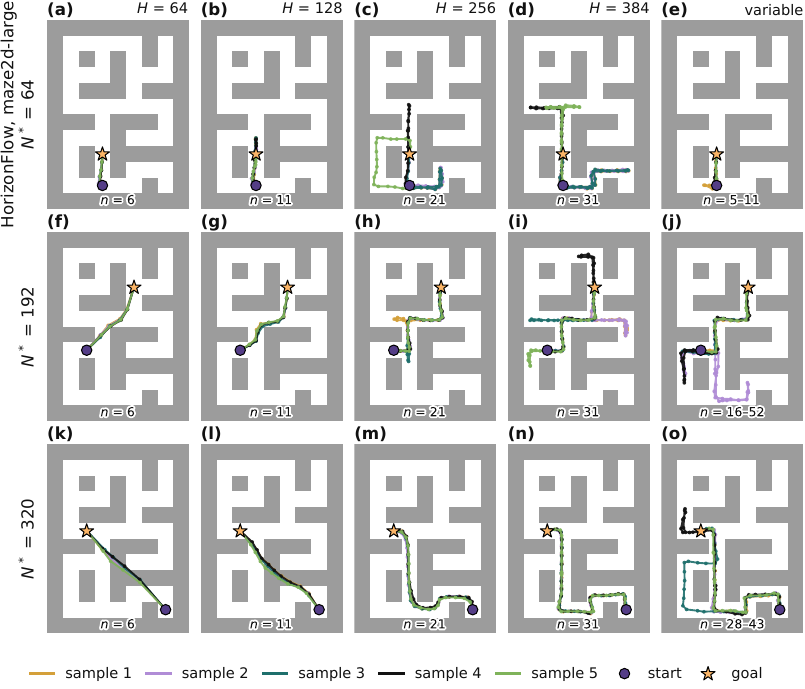}
  \caption{HorizonFlow XY plan samples on Maze2D-Large.
  Rows use the same start--goal pairs as
  \cfigref{fig:diffuser-plan-samples}; columns compare prescribed
  planning horizons with adaptive generation (variable).
  Colors distinguish generated reference plans, dots mark tokens,
  circles mark starts, stars mark goals, and gray regions are walls.
  Count labels in this figure include both anchors; ranges give
  the minimum and maximum counts among the displayed samples.}
  \label{fig:hf-xy-plan-samples}
\end{figure}

\paragraph{Qualitative plan samples.}
\label{app:horizon-diag:qualitative}
\Cref{fig:diffuser-plan-samples,fig:hf-xy-plan-samples} visualize
Diffuser and HorizonFlow XY reference plans for the same endpoints
under different length settings.
Short horizons can produce paths through walls for distant pairs,
whereas long horizons can introduce excursions beyond a nearby goal.
Each panel overlays five generated samples to show variation across
plans; this visualization is separate from the one-plan-per-pair-and-horizon
measurement protocol used in
\cfigref[(a--c)]{fig:diffuser-horizon-diagnostic}.
The XY examples also show variation in route shape and token count
under adaptive generation. These illustrative samples are not
executed trajectories or a quantitative comparison of success rates.

\subsection{Comparing Length Signals}
\label{app:horizon-diag:estimators}

We compare three signals on the same start--goal pairs: a value-derived
length, an explicit length prediction, and a generated plan count.
This diagnostic evaluates distance ranking without executing plans.

\paragraph{HIQL value-derived length.}
We use the official OGBench HIQL implementation with discount
$\gamma=0.995$, expectile 0.7, subgoal representation dimension 10,
batch size 1024, one million gradient steps, and seed 0.
Inputs are raw states $(x,y,\dot x,\dot y)$, and goals are drawn from
the training stream using OGBench's hierarchical goal-conditioned
sampler.
The target-network EMA, optimizer, and batch sampling follow the
official update procedure. We convert the value to a length using
\begin{equation}
  \hat n_{\mathrm{HIQL}}
  = \frac{\log\!\left(1+(1-\gamma)V(s,g)\right)}{\log\gamma}.
\end{equation}
The conversion is strictly decreasing on its valid domain, so
$\hat n_{\mathrm{HIQL}}$ and $-V$ induce the same ranking.

\paragraph{VHD length prediction.}
We use the scalar horizon $\hat L$ produced by our reproduced VHD
length predictor for each state--goal pair.

\paragraph{HorizonFlow count.}
We use the minimum non-anchor count among $K_{\mathrm{RP}}=64$
generated RP plans with FK steering.

\paragraph{Evaluation.}
For each threshold $x$, we compute the Spearman correlation
between each signal and $N^*$ over pairs satisfying $N^*\geq x$.
Increasing $x$ restricts the comparison to more distant goals,
testing whether the signals retain useful distance rankings at long
range. This measures agreement with nominal steps, not accuracy
against an optimal control horizon.

\begin{figure}[t]
  \centering
  \includegraphics[width=\linewidth]{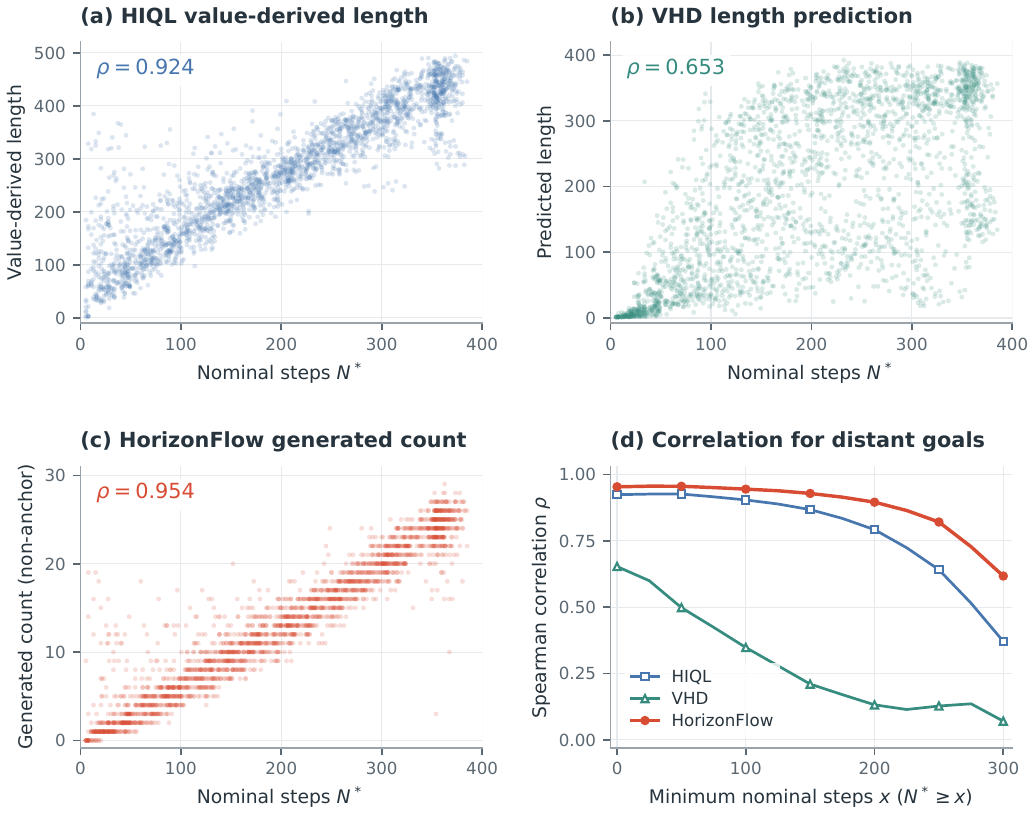}
  \caption{Length signals on matched Maze2D-Large start--goal pairs.
  (a--c) HIQL value-derived length, VHD predicted length, and HorizonFlow
  generated non-anchor count versus nominal steps $N^*$; $\rho$ denotes
  Spearman correlation. (d) Correlation recomputed on pairs with
  $N^*\geq x$ as the minimum nominal-step threshold $x$ increases.}
  \label{fig:length-signal-correlation}
\end{figure}

\paragraph{Results.}
\cfigref{fig:length-signal-correlation} reports 2,400 matched pairs,
with 300 pairs in each of eight nonempty 50-step bands of $N^*$.
Each method uses one trained model. VHD uses its unquantized prediction, and the
HorizonFlow count excludes the two fixed anchors in the stored plans.

HorizonFlow's generated count has the highest rank correlation with
nominal steps among the three signals in this diagnostic: $\rho=0.954$,
compared with $0.924$ for HIQL and $0.653$ for VHD.
The ordering persists on the 600 pairs with $N^*\geq300$, where the
correlations are $0.617$, $0.371$, and $0.070$, respectively.
Restricting the comparison changes both the target range and the
pair distribution. These cross-pair correlations assess distance
ranking; within-pair selection is evaluated in
\Cref{app:horizon-diag:within-pair}.

\section{Matched Full-System and Sampler Diagnostics}
\label{app:matched-diagnostics}

The diagnostics in \cref{app:horizon-diag} isolate geometric length signals
in Maze2D. Here we complement them with matched full-system controls on
AntMaze-Giant and HumanoidMaze-Giant, followed by implementation checks and
an integration-resolution sensitivity study. These experiments separate three
questions: whether route-planner (RP) performance survives an externally
predicted length, how action-prefix controller (PC) performance changes when its length
is fixed to the full horizon or the controller is replaced, and whether the practical sampler is sensitive to its
numerical resolution. Success rate is always measured from environment
rollouts; the non-rollout checks are identified explicitly.

\subsection{Route-Planner Length Assignment}
\label{app:rp-length-controls}

\begin{figure}[t]
\centering
\includegraphics[width=\linewidth]{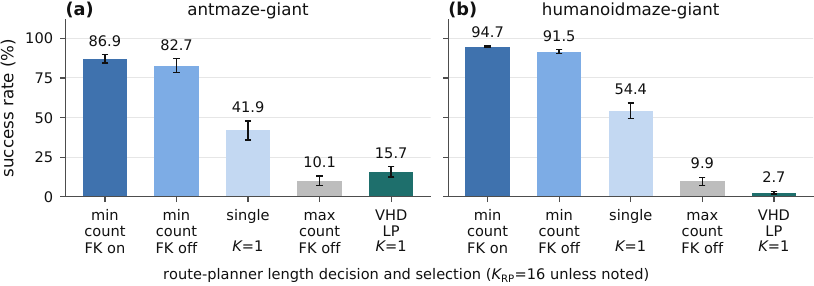}
\caption{\textbf{Full-system route-planner length controls.}
Success rates on AntMaze-Giant (left) and HumanoidMaze-Giant (right).
The bars compare minimum-count RP selection with FK enabled or disabled, a
single joint RP sample, maximum-count selection without FK, and a VHD-style
predicted-length control with one content sample. The RP uses $16$ candidates
unless $K=1$ is indicated. Error bars are standard deviations over three
training seeds, with $125$ evaluation problems per seed. The pretrained encoder
and PC are shared by the matched controls.}
\label{fig:review-rp-length}
\end{figure}

\paragraph{Matched setup.}
For the learned-length and externally assigned-length RPs in
\cfigref{fig:review-rp-length}, we match the Transformer size
($640$ hidden units, $10$ layers, and $8$ attention heads), dataset, stride,
length buckets ($8/16/32/64$), segment-length distribution, $100$k updates,
batch size $1024$, learning-rate schedule, and training seed. The
externally assigned-length model loads the encoder from the corresponding
HorizonFlow RP checkpoint and freezes it from the start of training. Thus,
both models use the same latent representation and can be paired with the
same pretrained HorizonFlow PC. The only training-objective change is that
the control RP contains all $m$ segment tokens from the outset: its two
endpoint anchors remain clean, the $m-2$ interior tokens share one flow time,
and the model is trained only on the velocity target, without birth or
completion targets. Training lengths are still sampled from the matched RP length
distribution; the length is fixed externally only at generation time.

The external length is provided by a VHD-style predictor trained with the
same architecture and losses as our VHD reproduction. It maps random Fourier
features of $(s,g)$ through a three-layer, width-$512$ MLP to
$f(s,g)\in[0,1]$, representing predicted steps divided by
$T_{\max}=63\Delta_{\mathrm{RP}}$. Its targets combine same-episode
anchor regression, bootstrapped dynamic-programming targets, consistency and
triangle constraints, and the boundary conditions $f(g,g)=0$ and $f\leq1$.
For this RP control, anchor offsets are
$\Delta_{\mathrm{RP}}\times\{1,2,4,8,16,32,63\}$, the batch size is $256$, and the
predictor is trained for $20$k updates with learning rate $3\times10^{-4}$.

\paragraph{Length conversion.}
We report the exact conversion used by the control:
\begin{equation}
  L=f(s,g)T_{\max}+1,\qquad
  m=\operatorname{clip}\!\left(
    \operatorname{round}\!\left(\frac{L}{\Delta_{\mathrm{RP}}}\right)+1,\,
    2,\,N_{\mathrm{RP}}
  \right).
  \label{eq:vhd-length-conversion}
\end{equation}
Here \texttt{round} is Python's rounding rule and the final $+1$ includes an
anchor. We call this a VHD-style learned-length control
rather than an oracle shortest-path assignment.

\paragraph{Results and scope.}
Minimum-count selection with FK reaches $86.9\%$ on AntMaze-Giant and
$94.7\%$ on HumanoidMaze-Giant; disabling FK changes these values to
$82.7\%$ and $91.5\%$. With selection removed, a single joint RP sample
reaches $41.9\%$ and $54.4\%$, whereas the VHD-style predicted-length control
with one content sample reaches $15.7\%$ and $2.7\%$. Selecting the
maximum-count candidate without FK reaches $10.1\%$ and $9.9\%$.

The $K=1$ comparison removes both best-of-$K$ selection and FK from the
contrast between joint generation and external length assignment. The control
plugs an external length module into the same system: the encoder, prefix
controller, data, length distribution, and training budget are shared, and
only the way the horizon is set differs. The comparison therefore isolates
horizon assignment within this system rather than benchmarking a separate
method. Conversely, the gap between one sample and minimum-count
selection shows that candidate generation and selection remain important
parts of the full RP recipe.

\subsection{Action-Prefix Controller Variants}
\label{app:pc-controls}

\begin{figure}[t]
\centering
\includegraphics[width=\linewidth]{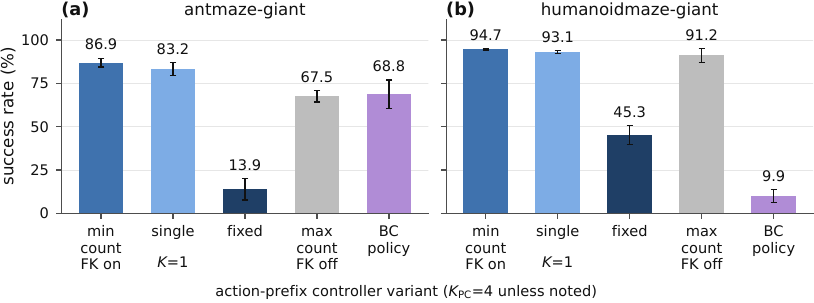}
\caption{\textbf{Action-prefix controller controls with the route planner
fixed.} Success rates on AntMaze-Giant (left) and HumanoidMaze-Giant (right)
for the default minimum-count PC, a single generated candidate, a full-horizon fixed-length Transformer, maximum-count selection without FK, and a
behavior-cloning policy. Each domain uses $125$ problems for each of seeds
$43$--$45$.}
\label{fig:review-pc}
\end{figure}

\paragraph{Setup.}
We next hold the default RP fixed and change only the PC. Each result in
\cfigref{fig:review-pc} aggregates $125$ problems for each of seeds
$43$--$45$, or $375$ rollouts per domain. The default PC generates four
FK-guided candidates and executes the minimum-count candidate. The
\emph{single} control removes candidate selection. The \emph{fixed} control is a full-horizon fixed-length PC: it uses the same
Transformer architecture and training distribution, is trained without
insertion, and always generates the full PC capacity $N_{\mathrm{PC}}$
(\ctabref{tab:domain-recipe}). The maximum-count
control selects the longest of four candidates with FK disabled. Finally, the
BC control replaces the generative PC with a behavior-cloning policy while
retaining the same RP.

\paragraph{Results and interpretation.}
The default PC reaches $86.9\%$ and $94.7\%$ on AntMaze-Giant and
HumanoidMaze-Giant, respectively. One candidate already reaches $83.2\%$ and
$93.1\%$, so the combined gain from four-candidate generation, FK, and
minimum-count selection is $3.7$ and $1.6$ percentage points. In contrast,
the full-horizon fixed-length Transformer reaches $13.9\%$ and $45.3\%$. This
full-horizon control keeps the architecture and training distribution fixed
and removes only insertion.
Maximum-count selection without FK reaches $67.5\%$ and $91.2\%$, while the
BC policy reaches $68.8\%$ and $9.9\%$.

The maximum-count condition reverses the count-guided inference rule
(ordering and FK together) and reduces success by $19.4$ and $3.5$ points.
The single-candidate result shows that the gain over the fixed-length and
BC controllers comes primarily from joint content--length generation
itself, with count-guided selection adding a further margin. HorizonFlow
executes only the first action of each generated prefix and then replans, so
the rollout success rate measures the accumulated consequence of those
repeated first-action decisions.

\subsection{Sampler Checks and Integration Resolution}
\label{app:sampler-diagnostics}

\begin{figure}[t]
\centering
\includegraphics[width=\linewidth]{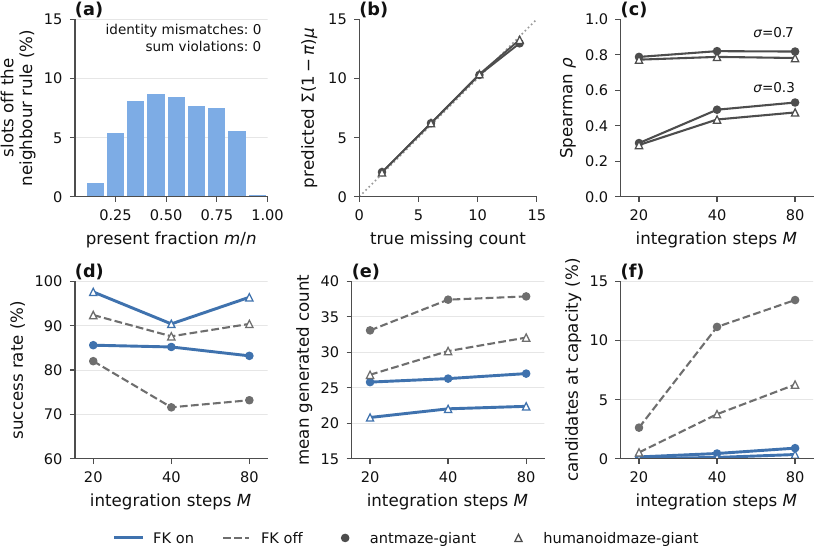}
\caption{\textbf{Sampler implementation checks and resolution sensitivity.}
Top: gap-target assignment and conservation checks, missing-count head
calibration, and rank correlation between intermediate FK scores and final
generated counts. Bottom: rollout success, average generated non-anchor count,
and fraction of RP candidates at the length capacity as the RP integration
resolution varies. Solid blue curves enable FK and dashed gray curves disable
it; circles denote AntMaze-Giant and triangles denote HumanoidMaze-Giant.}
\label{fig:review-sampler}
\end{figure}

\paragraph{Protocol.}
\cfigref{fig:review-sampler} uses one trained checkpoint (seed $43$) per
domain; its top row contains no environment rollouts. Panels (a) and
(b) use $20$k corruptions drawn from the training-data pipeline. Panel (a)
checks gap-target identities and total-count conservation. Panel (b) compares
the head's predicted missing count with the realized missing count across
noise levels. Panel (c) runs $K_{\mathrm{RP}}=16$ chains with FK disabled and
measures Spearman correlation between the intermediate FK score and each
chain's final count at $\sigma\in\{0.3,0.7\}$.

The bottom row evaluates $250$ problems per domain (five tasks for each of
evaluation seeds $0$--$49$), with budgets of $1000$ environment steps for
AntMaze-Giant and $4000$ for HumanoidMaze-Giant. We vary the RP integration
steps over $M\in\{20,40,80\}$. FK checkpoints scale proportionally:
$\{3,7\}$, $\{6,14\}$, and $\{12,28\}$. All conditions use
$K_{\mathrm{RP}}=16$ and minimum-count selection. The PC remains fixed at
$M=20$, $K_{\mathrm{PC}}=4$, with FK enabled.

\paragraph{Non-rollout checks.}
The gap-target test records zero identity mismatches and zero total-count
violations over the $20$k sampled corruptions. The nonzero bars in panel~(a) report the fraction of slots outside the displayed neighbour-rule case; the identity and sum checks are the annotated zero-valued diagnostics. The missing-count predictions
track the diagonal across noise levels, with aggregate signed biases of
$+0.01$ on AntMaze-Giant and $+0.04$ on HumanoidMaze-Giant. This validates
the count head on the sampled training distribution; it is not a calibration
claim for every output of the denoiser. At $\sigma=0.3$, intermediate-score
rank correlations are approximately $0.3$--$0.5$; at $\sigma=0.7$, they rise
to about $0.8$. Accordingly, FK uses this quantity for within-call ranking,
not as a calibrated estimate of the eventual count.

\paragraph{Resolution sensitivity.}
With FK enabled, AntMaze-Giant success changes from approximately
$86\%$ to $85\%$ and $83\%$ as $M$ increases, while HumanoidMaze-Giant
changes from approximately $98\%$ to $90\%$ and $96\%$. The intermediate
HumanoidMaze-Giant dip therefore does not continue at $M=80$. Average
generated counts move by only one to two tokens, and fewer than $1\%$ of
candidates reach the length capacity. Without FK, generated plans become
longer as $M$ increases; on AntMaze-Giant the average count grows from about
$33$ to $38$, and the capacity fraction grows from $2.6\%$ to $13.4\%$.
The corresponding HumanoidMaze-Giant capacity fraction grows from $0.5\%$ to
$6.3\%$. This drift without steering is one reason FK is part of the
default recipe.

\section{Count-Based Steering Score from the Hurdle--ZTP Head}
\label{app:count-derivation}

At a steering checkpoint, let $G(x_\sigma)$ be the gaps between the tokens
currently present in particle $x_\sigma$. For each gap $g$, let $K_g$ denote
the gapwise missing-count random variable represented by the
hurdle--ZTP head at that checkpoint. Its distribution is
\begin{equation}
  \Pr(K_g=0\mid x_\sigma)=\pi_\theta(g\mid x_\sigma)
  \label{eq:hurdle-zero}
\end{equation}
and, for $k\geq1$,
\begin{equation}
  \Pr(K_g=k\mid K_g>0,x_\sigma)
  =
  \frac{e^{-\lambda_\theta(g\mid x_\sigma)}
        \lambda_\theta(g\mid x_\sigma)^k}
       {k!\left(1-e^{-\lambda_\theta(g\mid x_\sigma)}\right)}.
  \label{eq:ztp-pmf}
\end{equation}
The conditional mean of a zero-truncated Poisson variable is
\begin{equation}
  \mathbb{E}[K_g\mid K_g>0,x_\sigma]
  =
  \frac{\lambda_\theta(g\mid x_\sigma)}
       {1-e^{-\lambda_\theta(g\mid x_\sigma)}}.
  \label{eq:ztp-mean}
\end{equation}
Applying the law of total expectation to the hurdle event gives
\begin{equation}
  \mathbb{E}[K_g\mid x_\sigma]
  =
  \bigl(1-\pi_\theta(g\mid x_\sigma)\bigr)
  \frac{\lambda_\theta(g\mid x_\sigma)}
       {1-e^{-\lambda_\theta(g\mid x_\sigma)}}.
  \label{eq:hurdle-ztp-mean}
\end{equation}
Excluding the fixed start and goal anchors, we construct the
count-based steering score by adding the current token count to
the expected missing counts under the predictive heads:
\begin{equation}
  \begin{aligned}
  \hat n(x_\sigma)
  &:= |x_\sigma| + \sum_{g\in G(x_\sigma)}
      \mathbb{E}_\theta[K_g\mid x_\sigma] \\
  &= |x_\sigma|
     + \sum_{g\in G(x_\sigma)}
       \bigl(1-\pi_\theta(g\mid x_\sigma)\bigr)
       \frac{\lambda_\theta(g\mid x_\sigma)}
            {1-e^{-\lambda_\theta(g\mid x_\sigma)}}.
  \end{aligned}
  \label{eq:final-count-derivation}
\end{equation}
This gives Equation~\ref{eq:nhat}. The expectations are taken under
the gapwise hurdle--ZTP predictive heads; their sum does not require
the $K_g$ to be independent. We use the resulting score as a
steering potential, rather than identifying it with the conditional
expected final count of the practical sampler. No rollout reward or
separately learned value enters its construction.

\section{Method Details and Pseudocode}
\label{app:method}

We detail the token representation, training procedure, sampler, and
control loop used for the OGBench navigation experiments.
\Cref{app:method:common} specifies the shared recipe and its
Maze2D and visual-manipulation settings.
Algorithms~\ref{alg:corrupt}--\ref{alg:control} summarize the procedures;
\ctabref{tab:recipe} records the default recipe. We retain the
main-text notation: $r$ is the order coordinate, $\sigma$ is the global generation
clock, $t_i$ is a token's local refinement time, and $h$ indexes
environment steps.

\subsection{Tokens, Storage, and Architecture}
\label{app:method:objects}
\label{app:method:model}

\paragraph{Generated tokens and conditioning anchors.}
Each generated token is $x_i=(r_i,c_i)$, with continuous content $c_i$.
RP generates tokens $(r,z)$ with $z\in\mathbb{R}^{16}$; PC generates
$(r,a)$, with one action per interior token. PC storage also contains
conditioning fields for state information, but these fields are not
read from or predicted for interior tokens. The start anchor carries
the current proprioceptive state; the goal anchor carries the frozen
RP encoder's latent of the goal frame, padded into the conditioning
field. Their action fields are zero. Thus the storage layout does
not make PC a state--action trajectory generator.

Both planners have two fixed anchors at $r=-1$ and $r=+1$, excluded
from the tokenwise FM loss and generated count. We write $k$ for the
active token count including these anchors and $n=k-2$ for the generated
count. A clean sequence has coordinates $r_i=2(i-1)/(m-1)-1$, where
$m$ includes anchors. Gaussian noise and intermediate flow states
need not have coordinates in $[-1,1]$; the anchor values remain fixed and
the other tokens are ordered by their current $r$ coordinates.

\paragraph{Padded storage and gaps.}
Token sets are stored in fixed-capacity arrays with padding masks:
$N_{\mathrm{RP}}=64$ and $N_{\mathrm{PC}}=32$, including anchors.
We denote a generic storage capacity by $C_{\mathrm{store}}=N+2$,
where $N$ is the non-anchor token capacity in the main-text plan space;
$C_{\mathrm{store}}$ equals $N_{\mathrm{RP}}$ or $N_{\mathrm{PC}}$ at the respective level.
Uppercase $X$ denotes anchor-inclusive storage, whereas lowercase $x$
denotes the generated non-anchor plan. Thus $|x|=n=k-2$.
Only the $k$ active rows are processed as plan tokens; padding does
not count toward plan length. A learned beginning-of-sequence (BOS)
token is prepended to the network input, without a local-time embedding.
It is distinct from the two conditioning anchors and is not counted
in $k$ or the generated count. The count and completion heads produce
outputs at the BOS and active-token positions, giving $k+1$ gap outputs,
including the slots before and after the active tokens. Boundary
gaps are included in the insertion supervision, even though clean
sequences have no missing targets outside the endpoint anchors.

\paragraph{Transformer and output heads.}
Both planners use pre-layer-normalized Transformer blocks with eight
attention heads and additive sinusoidal embeddings of each token's
local refinement time. The network operates on the noisy plan $x_\sigma$ defined in
\Cref{sec:method-train}. The RP model has width 640 and ten blocks (51.3M
parameters), with a linear input projection for $(r,z)$. The PC model
has width 512 and four blocks (14.0M parameters). Its state-free input
uses separate projections for interior $(r,a)$ tokens and anchor
conditioning, plus a type embedding. Its velocity output covers only
$(r,a)$, never intermediate states.

The three heads predict token velocities $v_\theta$, positive-count
parameters $\lambda_\theta$ through softplus, and gap-completion
probabilities $\pi_\theta$ through sigmoid. The parameter $\lambda$
is the base Poisson parameter of the positive-count distribution;
its conditional mean is $\lambda/(1-e^{-\lambda})$, not $\lambda$.

\subsection{Training}
\label{app:training-details}
\label{app:method:train}

\paragraph{Segments and variable lengths.}
Training examples are sampled from individual dataset episodes.
The clean count $m$, including anchors, is sampled log-uniformly
from $[2,64]$ for RP and $[3,32]$ for PC. Examples are bucketed at
capacities $\{8,16,32,64\}$ and $\{8,16,32\}$, respectively, with
bucket allocations proportional to the corresponding probability
mass, up to integer batch-size rounding. Specifically, sampling a
uniform real value on $[\log m_{\min},\log(m_{\max}+1))$ and
exponentiating and flooring it gives
$p(m)=\log((m+1)/m)/\log((m_{\max}+1)/m_{\min})$.
Each bucket samples its disjoint count interval under this law;
rounded example allocations are adjusted in the largest bucket to
preserve the total batch size. RP interiors are sampled at stride
$\Delta_{\mathrm{RP}}$, whereas PC interiors use unit-stride actions.
For PC, the action-prefix length is $n=m-2$; these interior tokens represent
$a_h,\ldots,a_{h+n-1}$ from a segment starting at environment step $h$.
Interior token $q\in\{1,\ldots,n\}$ carries $a_{h+q-1}$, the logged
action leading into frame $s_{h+q}$. These frame indices specify the
training-data correspondence; PC generates actions, not interior states.
Anchor action fields are set to zero.

\paragraph{Local-prefix targets.}
\label{app:lo-prefix}
For a sampled action prefix of $n$ environment steps, the hindsight
goal is taken at offset
\begin{equation}
  \ell_{\mathrm{goal}}=K_{\mathrm{div}}n.
  \label{eq:lo-goal-distance}
\end{equation}
The goal anchor contains the frozen RP encoder's latent of that
frame. With $K_{\mathrm{div}}=1$, the last interior token and goal
anchor correspond to the same frame $s_{h+n}$: the former carries
the final action leading into that frame, while the latter provides
goal conditioning. Larger divisors pair the same action prefix with
a more distant goal frame. At fixed $K_{\mathrm{div}}>0$, the target count $n$ and goal offset $K_{\mathrm{div}}n$ have the same ordering. This training relation motivates using generated count as a temporal proxy when ranking prefixes, including when $K_{\mathrm{div}}>1$. The default controller executes the selected prefix's first action and replans; it does not commit to executing the whole prefix or treat its count as a verified arrival time. In particular, PC training lengths
are sampled rather than fixed at
$\Delta_{\mathrm{RP}}/K_{\mathrm{div}}$. The default divisor is one for PointMaze and
AntMaze, two for HumanoidMaze-Medium/Large, and four for
HumanoidMaze-Giant. It is distinct from the candidate count $K$.

\paragraph{Interleaved corruption.}
\label{app:corruption}
We sample $\sigma\sim\mathcal U(0,2)$ and independent offsets
$u_i\sim\mathcal U(0,1)$. A non-anchor token is present when
$\sigma-u_i\geq0$ and has local time
$t_i=\operatorname{clip}_{[0,1]}(\sigma-u_i)$; anchors are always
present and clean at $t_i=1$. Deleted tokens supply the gapwise
missing-count targets. 
After interpolation, the model receives tokens sorted by their
noised order coordinates. Algorithm~\ref{alg:corrupt} summarizes
this corruption. For $\sigma<1$, insertions and refinement coexist;
for $\sigma\geq1$, all clean tokens are present and refinement
continues, mirroring the sampler's two-phase clock schedule.

\begin{algorithm}[t]
\caption{\textsc{Corrupt}: interleaved corruption of a clean plan}
\label{alg:corrupt}
\begin{algorithmic}[1]
\Require Clean tokens $X_1=(x_{1,i})_{i=1}^{m}$ in route order; anchor mask $A$
\State Sample $\sigma\sim\mathcal U(0,2)$ and $u_i\sim\mathcal U(0,1)$ independently
\State Form present-token mask $P$, local times $\mathbf t$, and gap counts $\{c_g\}$ as defined above
\State $k\gets\sum_i P_i$ \Comment{Number of present tokens; $k+1$ gaps include boundaries}
\State Draw noise $\epsilon$ using the noise distribution
\State $x_{\mathbf t,i}\gets(1-t_i)\epsilon_i+t_i x_{1,i}$ for present tokens
\State $v_i^\star\gets x_{1,i}-\epsilon_i$ on supervised, non-anchor coordinates
\State Sort present tokens by noised $r$, carrying times, targets, and masks
\State \Return Corrupted plan, local times, velocity targets, anchor mask, and $\{c_g\}$
\end{algorithmic}
\end{algorithm}

\paragraph{Flow-matching and insertion losses.}
Let $\mathcal S$ contain all present non-anchor tokens in one
device-local bucket batch, and let
$D$ denote the number of supervised coordinates: $1+d_z$ for RP
and $1+d_a$ for PC. The velocity loss is
\begin{equation}
  \mathcal L_{\mathrm{FM}}
  =\mathbb E\!\left[
    \frac{1}{D|\mathcal S|}\sum_{i\in\mathcal S}
    \omega(t_i)\left\|v_\theta(X_{\mathbf t},\mathbf t)_i-v_i^\star\right\|_2^2
  \right],
  \label{eq:fm-loss-detail}
\end{equation}
with zero token loss when $\mathcal S$ is empty. Anchors and padding
are masked. The logit-normal importance weight is proportional to
\begin{equation}
  \omega(t)\propto
  \frac{\exp\!\left[-\tfrac12\left(\log\tfrac{t}{1-t}\right)^2\right]}
       {t(1-t)},
  \qquad 0<t<1,
\end{equation}
evaluated after clipping $t$ to $[10^{-3},1-10^{-3}]$ and
normalized to mean one over $\mathcal S$~\citep{esser2024sd3}.
Thus the reduction pools supervised tokens within a bucket batch;
it is not an unweighted average of per-plan velocity losses.
For $c_g$ missing clean tokens in gap $g$, the insertion objective is
\begin{equation}
  \begin{aligned}
  \mathcal L_{\mathrm{ins}}
  &=\mathbb E\!\left[\frac{1}{|G|}\sum_{g\in G}\ell_g\right],\\
  \ell_g
  &=\mathrm{BCE}\!\left(\pi_\theta(g),\mathbf1[c_g=0]\right)\\
  &\quad+\mathbf1[c_g>0]\left(
    \lambda_\theta(g)-c_g\log\lambda_\theta(g)
    +\log(1-e^{-\lambda_\theta(g)})\right).
  \end{aligned}
  \label{eq:ins-loss-detail}
\end{equation}
Here $|G|=k+1$, including the boundary gap outputs. The
parameter-independent factorial term is omitted. Each example
first averages over its valid gaps, and these insertion losses are
then averaged over the examples in the bucket batch.

Across buckets, RP weights the full loss by the fraction of examples
in each bucket. PC weights the FM term by the fraction of supervised
tokens in each bucket and the insertion term by the fraction of
examples. Gradients are averaged across devices before the optimizer
update.

\paragraph{Route planner's latent encoder and gradient routing.}
\label{app:method:hi}
The encoder is a two-layer MLP with a 64-unit hidden layer,
\begin{equation}
  z=E_\phi(s)=\tanh\!\left(W_2\,\operatorname{SiLU}(W_1s)\right)
  \in\mathbb R^{16}.
\end{equation}
It is trained jointly with the route planner, but receives gradients
only from insertion supervision and temporal contrastive
regularization. FM gradients are blocked at the encoder so that its
representations cannot shrink merely to simplify the velocity target.
For temporally nearby positive states and in-batch negatives, we use
\begin{equation}
  \mathcal L_{\mathrm{NCE}}
  =\mathbb E\!\left[-\log
  \frac{\exp(\bar z_j^\top\bar z_{j+\Delta}/\kappa)}
       {\sum_{j'}\exp(\bar z_j^\top\bar z_{j'+\Delta'}/\kappa)}
  \right],
  \label{eq:nce-loss-detail}
\end{equation}
where $\bar z$ is $\ell_2$-normalized,
$\Delta\sim\mathcal U\{1,\ldots,4\}$, and $\kappa=0.2$.
The offset is measured in subsampled RP frame slots, corresponding
to $\Delta\Delta_{\mathrm{RP}}$ environment steps before clipping at
the episode boundary. For a bucket of capacity $C$, the first slot
is sampled from $\{0,\ldots,C-5\}$ and the positive is $\Delta$
slots later. These pairs are sampled from the full frame buffer,
including frames beyond the example's sampled clean count; they
are not restricted to tokens present after corruption. Negatives
are the positive frames of other examples in the same device-local
bucket batch.
This encourages temporal structure and separation from negatives;
it does not guarantee an injective representation.
RP uses $\mathcal L_{\mathrm{FM}}+\mathcal L_{\mathrm{ins}}
+0.1\mathcal L_{\mathrm{NCE}}$. PC uses only
$\mathcal L_{\mathrm{FM}}+\mathcal L_{\mathrm{ins}}$, with the RP
encoder frozen for goal conditioning. At test time, the same encoder
forms both RP anchors, and generated subgoal latents are passed to PC
directly.

\begin{algorithm}[t]
\caption{One training step}
\label{alg:train}
\begin{algorithmic}[1]
\Require Component (RP or PC), dataset, parameters $\theta$, and RP encoder $E_\phi$
\State Sample a bucketed batch of episode segments and construct clean tokens
\State For RP, encode subgoal states; for PC, encode goal frames with frozen $E_\phi$
\State Apply \textsc{Corrupt} to each example (Algorithm~\ref{alg:corrupt})
\State Evaluate the velocity, positive-count, and completion heads with padding masks
\State $\mathcal L\gets\mathcal L_{\mathrm{FM}}+\mathcal L_{\mathrm{ins}}$
\If{RP training}
  \State $\mathcal L\gets\mathcal L+0.1\mathcal L_{\mathrm{NCE}}$
  \State $g_\phi\gets\nabla_\phi(\mathcal L_{\mathrm{ins}}+0.1\mathcal L_{\mathrm{NCE}})$
\EndIf
\State $g_\theta\gets\nabla_\theta\mathcal L$
\State Clip gradient norm to 1.0 and update trainable parameters with AdamW
\State Apply the cosine learning-rate schedule and update parameter EMA (0.999)
\end{algorithmic}
\end{algorithm}

\subsection{Sampling and Candidate Selection}
\label{app:method:sample}

\paragraph{Two-phase rollout.}
At each step, the model evaluates the partial plan, refines existing
tokens, determines how many new tokens to initialize, and sorts the
updated token set. Algorithm~\ref{alg:advance} summarizes this update;
the finite-step implementation is specified below.
Sampling advances $\sigma$ from zero to two in $M=20$ Euler steps
of size $\delta=2/M=0.1$. Each token has its own refinement time
$t_i$, and both anchors start at $t_i=1$. Insertions are enabled
only for $\sigma<1$. After that cutoff the token count is fixed,
while the remaining steps finish transporting the last-born tokens.
The sampler uses $\sigma$ to determine the time-dependent birth
probabilities, the effective birth-step size $\min(\delta,1-\sigma)$
before the cutoff, the insertion cutoff, and the FK checkpoint times.
The sampler uses EMA parameters.

\paragraph{Insertion probability.}
For a gap predicted at the current state, let
$\mu_g=\lambda_g/(1-e^{-\lambda_g})$ be the expected missing count
conditional on the gap being incomplete. Motivated by the uniform
reveal-time schedule used in training, we use the model-based insertion
intensity $(1-\pi_g)\mu_g/(1-\sigma)$ for $\sigma<1$ and the finite-step rule
\begin{equation}
  p_g^{\mathrm{birth}}(\sigma)=
  \begin{cases}
  (1-\pi_g)\operatorname{clip}_{[0,1]}\!\left(
    \dfrac{\min(\delta,1-\sigma)\mu_g}{1-\sigma}\right),
    &\sigma<1,\\
  0,&\sigma\geq1.
  \end{cases}
  \label{eq:birth-probability}
\end{equation}
The same hurdle--ZTP mean enters
both the insertion intensity and the count-based steering score in
Equation~\ref{eq:nhat}.
For numerical stability, $\mu_g$ can be evaluated using
$-\operatorname{expm1}(-\lambda_g)$ in the denominator and its limit
$\mu_g\to1$ as $\lambda_g\to0$.

\paragraph{Finite-step implementation.}
\label{app:method:sample-implementation}
We implement Equation~\ref{eq:birth-probability} with two independent
uniform tests during the insertion phase. A birth is proposed only if
both tests pass: the first with probability $1-\pi_g$ and the second
with the clipped time-scaled conditional mean in
Equation~\ref{eq:birth-probability}. Thus the completion gate
$1-\pi_g$ multiplies the clipped factor.
Each gap proposes at most one birth per step; the proposals are pooled,
and their total is capped by the remaining storage capacity.
Newborn tokens, including their order coordinates, are initialized from
standard Gaussian noise at local time zero, as in training corruption;
subsequent flow refinement of $r$ places them along the plan.

\begin{algorithm}[t]
\caption{\textsc{Advance}: one insertion--flow sampling step}
\label{alg:advance}
\begin{algorithmic}[1]
\Require Active tokens $X$, local times $\mathbf t$, anchor mask $A$, clock $\sigma$, step $\delta$, storage capacity $C_{\mathrm{store}}$
\State $(v,\lambda,\pi)\gets f_\theta(X_\sigma)$ \Comment{$X_\sigma=(X,\mathbf t,A)$, the noisy plan at clock $\sigma$}
\State $\Delta t_i\gets\min(\delta,1-t_i)$ for non-anchors; zero for anchors
\State $x_i\gets x_i+v_i\Delta t_i$ on generated coordinates; $t_i\gets t_i+\Delta t_i$
\If{$\sigma<1$}
  \State Determine the number of new tokens from $(\lambda,\pi)$ using Equation~\ref{eq:birth-probability} and the available capacity
  \State Initialize and append these tokens with local time zero (\Cref{app:method:sample-implementation})
\EndIf
\State Sort active tokens by $r$, carrying local times and anchor identities
\State \Return Updated $X$, $\mathbf t$, and $A$
\end{algorithmic}
\end{algorithm}

\paragraph{Best-of-$K$.}
Without FK, candidates use independent random draws from the same
anchors. We select the minimum generated count as in
Equation~\ref{eq:minn}; including the two anchors would give the same
ranking because every candidate contains both. In particular, $K=1$
uses a single unranked sample. The rule favors compact generated
plans, not a certified shortest feasible trajectory. Its usefulness
is evaluated through the distance-correlation and executed-control
experiments rather than inferred solely from the training objective. For $K=1$, final selection is trivial and the normalized FK weight is one, so resampling returns the sole candidate; FK on and off therefore give the same single-candidate sampling distribution. Without steering, let $X_1,\ldots,X_K$ be independent samples from the same conditional generator and let $J$ be an independent uniform index. Then $X_J$ has the same conditional distribution as $X_1$, making the single-candidate baseline a random-selection reference in expectation. This equivalence does not generally apply to a multi-candidate pool modified by FK. In a per-level sweep, the other level retains its specified configuration.

\paragraph{Feynman--Kac steering.}
\label{app:method:fk}
The default control recipe uses FK at both levels, with checkpoints after
Euler steps $\mathcal C=\{3,7\}$ and $\beta=0.5$.
At each checkpoint, we evaluate the count-based steering score in
Equation~\ref{eq:nhat}, derived in \Cref{app:count-derivation}.
This score combines the current generated count with the predictive
heads' expected missing counts; it is used to favor lower-count
candidates, not as an exact conditional expectation of the practical
sampler's final count.
For candidate $j$, write $\hat n_j=\hat n(x_\sigma^{(j)})$. We
stabilize the resampling weights numerically as
\begin{equation}
  w_j\propto\exp\!\left[-\beta
    \left(\hat n_j-\min_{j'}\hat n_{j'}\right)\right].
\end{equation}
Subtracting the minimum changes neither the normalized weights nor
the resampling distribution. Duplicated chains receive independent
random keys for subsequent stochastic operations, not additional
perturbations to already generated tokens.
The final decision remains minimum-count selection.
\cfigref[(a,~b)]{fig:compute-alloc} compares FK-on and FK-off variants
at the swept level while retaining FK at the other level.

\begin{algorithm}[t]
\caption{\textsc{Plan}: FK steering and minimum-count selection at either level}
\label{alg:fk}
\begin{algorithmic}[1]
\Require Conditioning anchors (RP or PC), $K$, storage capacity $C_{\mathrm{store}}$, steps $M$, checkpoints $\mathcal C$, strength $\beta$
\State Initialize $K$ two-anchor chains with local times one and independent random keys
\State $\delta\gets2/M$
\For{$q=0,\ldots,M-1$}
  \State Advance each chain using \textsc{Advance} at $\sigma=q\delta$
  \If{$q+1\in\mathcal C$}
    \State Evaluate each updated chain and compute the steering score $\hat n_j$ from Equation~\ref{eq:nhat}
    \State Normalize $w_j\propto\exp[-\beta(\hat n_j-\min_{j'}\hat n_{j'})]$
    \State Resample $K$ full chain states with replacement from $\operatorname{Cat}(w)$
    \State Assign independent future random keys to duplicated chains
  \EndIf
\EndFor
\State \Return Non-anchor tokens of the chain with minimum generated count
\end{algorithmic}
\end{algorithm}

\subsection{Hierarchical Control}
\label{app:method:control}

The default controller holds the first generated RP subgoal for
$H_{\mathrm{RP}}=\Delta_{\mathrm{RP}}/2$ environment steps. It does
not advance a pointer along the route or switch targets on proximity.
If RP generates no interior token, the target is the final-goal latent.
PC replans at every environment step and executes only its first
generated action from the selected candidate, clipped to $[-1,1]$; an empty PC plan yields a zero
action. \cfigref[(c,~d)]{fig:compute-alloc} varies the RP holding
period and the fraction of a PC prefix executed per update.

\begin{algorithm}[t]
\caption{HorizonFlow at test time: default OGBench control loop}
\label{alg:control}
\begin{algorithmic}[1]
\Require Goal state $g$, period $H_{\mathrm{RP}}$, $K_{\mathrm{RP}}=16$, $K_{\mathrm{PC}}=4$
\State Observe $s_0$ and set $z_g\gets E_\phi(g)$
\For{$h=0,\ldots,T-1$}
  \If{$h\bmod H_{\mathrm{RP}}=0$}
    \State $W\gets\textsc{Plan}((-1,E_\phi(s_h)),(+1,z_g),K_{\mathrm{RP}},N_{\mathrm{RP}},M,\mathcal C,\beta)$
    \State $z^{\mathrm{tgt}}\gets$ first latent in $W$, or $z_g$ if $W$ is empty
  \EndIf
  \State Construct PC anchors from $(-1,s_h)$ and $(+1,z^{\mathrm{tgt}})$, with zero action fields
  \State $P\gets\textsc{Plan}$ from these anchors using $K_{\mathrm{PC}}$, $N_{\mathrm{PC}}$, $M$, $\mathcal C$, and $\beta$
  \State $a_h\gets$ first action in $P$, or $\mathbf0$ if $P$ is empty
  \State Execute $\operatorname{clip}_{[-1,1]}(a_h)$ and observe $s_{h+1}$
\EndFor
\end{algorithmic}
\end{algorithm}

\subsection{Default OGBench Recipe}
\label{app:method:recipe}

The recipe below applies to the nine OGBench navigation environments.
The stride, corresponding RP update period, and PC target divisor
vary by environment group; the architecture and optimization settings
are shared. These defaults do not override explicitly swept ablations
or the $K_{\mathrm{RP}}=64$ horizon-estimator diagnostic.
Both learning-rate schedules warm up linearly from zero for 2,000
updates, then decay with a cosine schedule to 2\% of the peak
learning rate. Weight decay is decoupled through AdamW.

\begin{table}[t]
\centering
\small
\setlength{\tabcolsep}{4pt}
\renewcommand{\arraystretch}{1.12}
\caption{Default OGBench architecture, training, and inference settings.}
\label{tab:recipe}
\begin{tabular}{@{}p{0.24\linewidth}p{0.35\linewidth}p{0.35\linewidth}@{}}
\toprule
Setting & Route planner (RP) & Prefix controller (PC)\\
\midrule
Generated token & $(r,z)$, $d_z=16$ & $(r,a)$; state-free interiors\\
Capacity (anchors included) & 64 & 32\\
Clean count distribution & Log-uniform $[2,64]$ & Log-uniform $[3,32]$\\
Bucket capacities & 8, 16, 32, 64 & 8, 16, 32\\
Transformer width / blocks & 640 / 10 & 512 / 4\\
Attention heads & 8 & 8\\
Parameters & 51.3M & 14.0M\\
Encoder & 64-unit hidden layer; 16-D output & Frozen RP encoder for goal conditioning\\
Loss & FM + insertion + $0.1$ NCE & FM on $(r,a)$ + insertion\\
Optimizer & AdamW & AdamW\\
Learning rate & $6\times10^{-4}$ & $8\times10^{-4}$\\
Schedule / weight decay & Cosine / 0.01 & Cosine / 0.01\\
Gradient clipping / EMA & 1.0 / 0.999 & 1.0 / 0.999\\
Batch size / updates & 1024 / 100k & 1024 / 1M\\
Sampler & 20 Euler steps on $[0,2]$ & 20 Euler steps on $[0,2]$\\
Candidate count & $K_{\mathrm{RP}}=16$ & $K_{\mathrm{PC}}=4$\\
FK steering & Steps 3, 7; $\beta=0.5$ & Steps 3, 7; $\beta=0.5$\\
Control & First subgoal, updated every $\Delta_{\mathrm{RP}}/2$ steps & Replan each step; execute first action\\
\bottomrule
\end{tabular}
\medskip
\begin{tabular}{@{}lccc@{}}
\toprule
Environment group & $\Delta_{\mathrm{RP}}$ & $H_{\mathrm{RP}}$ & $K_{\mathrm{div}}$\\
\midrule
PointMaze / AntMaze & 16 & 8 & 1\\
HumanoidMaze-Medium / Large & 32 & 16 & 2\\
HumanoidMaze-Giant & 64 & 32 & 4\\
\bottomrule
\end{tabular}
\end{table}

\subsection{Common Recipe across Benchmark Domains}
\label{app:method:common}

\paragraph{Shared architecture and optimization.}
The architecture, losses, and optimization schedule in
\ctabref{tab:recipe} also apply to Maze2D and visual manipulation.
RP's contrastive loss uses temperature 0.2 and positive offsets of
one to four subsampled frames. The temporal spacing, observation
encoding, and batch size vary as described below.

\paragraph{Episode-length rule.}
Let $T$ be the environment's maximum episode length in steps.
With RP storage capacity $N_{\mathrm{RP}}=64$, including anchors,
we set
\begin{equation}
 \Delta_{\mathrm{RP}}=\left\lceil\frac{T}{N_{\mathrm{RP}}-1}\right\rceil,
 \qquad
 H_{\mathrm{RP}}=\max\!\left(1,\left\lfloor\frac{\Delta_{\mathrm{RP}}}{2}\right\rfloor\right),
 \qquad
 K_{\mathrm{div}}=\max\!\left(1,\left\lfloor\frac{\Delta_{\mathrm{RP}}}{16}\right\rfloor\right).
 \label{eq:common-recipe}
\end{equation}
The denominator counts intervals between the 64 stored positions.
The corresponding PC capacity is
$N_{\mathrm{PC}}=\min(32,\lfloor2\Delta_{\mathrm{RP}}/K_{\mathrm{div}}\rfloor)$.
These are storage and training limits; they do not prescribe the
number of non-anchor tokens generated in an individual plan.
\ctabref{tab:domain-recipe} lists the resulting settings.
Maze2D and Multi2D use the same trained models and recipe; their
evaluation goal protocols differ.

\begin{table}[t]
\centering
\small
\setlength{\tabcolsep}{5pt}
\caption{Environment-dependent settings of the common recipe.
$T$ is the episode limit, $\Delta_{\mathrm{RP}}$ the training stride,
$H_{\mathrm{RP}}$ the RP update period, and $N_{\mathrm{PC}}$ the PC
capacity including anchors. Batch size applies to both stages.}
\label{tab:domain-recipe}
\begin{tabular}{@{}lrrrrrr@{}}
\toprule
Environment & $T$ & $\Delta_{\mathrm{RP}}$ & $H_{\mathrm{RP}}$ & $K_{\mathrm{div}}$ & $N_{\mathrm{PC}}$ & Batch\\
\midrule
Maze2D-UMaze & 300 & 5 & 2 & 1 & 10 & 1024\\
Maze2D-Medium & 600 & 10 & 5 & 1 & 20 & 1024\\
Maze2D-Large & 800 & 13 & 6 & 1 & 26 & 1024\\
PointMaze / AntMaze & 1000 & 16 & 8 & 1 & 32 & 1024\\
HumanoidMaze-Medium / Large & 2000 & 32 & 16 & 2 & 32 & 1024\\
HumanoidMaze-Giant & 4000 & 64 & 32 & 4 & 32 & 1024\\
Visual Cube-Single & 200 & 4 & 2 & 1 & 8 & 256\\
Visual Cube-Double & 500 & 8 & 4 & 1 & 16 & 256\\
Visual Cube-Triple & 1000 & 16 & 8 & 1 & 32 & 256\\
Visual Scene & 750 & 12 & 6 & 1 & 24 & 256\\
\bottomrule
\end{tabular}
\end{table}

\paragraph{Observation encoding and batching.}
Maze2D and navigation use the state encoder described above.
For visual manipulation, RP instead uses an IMPALA-style image
encoder with a 16-dimensional tanh output. PC encodes the current
image with its own trainable image encoder into 128 features for
the start anchor; its goal anchor uses the frozen RP latent.
PC interiors still contain only order coordinates and actions. Thus the
proprioceptive start-anchor description above applies to state-input
tasks, and the visual models have additional image-encoder parameters.
The visual batch size is 256, compared with 1024 for state-input tasks.

Clean counts follow the same log-uniform law, with bounds
$[2,64]$ for RP and $[3,N_{\mathrm{PC}}]$ for PC.
State-input RP uses buckets $\{8,16,32,64\}$; visual RP uses a
single padded capacity of 64. State-input PC uses the members of
$\{8,16,32\}$ strictly below its capacity, followed by the capacity
itself. Visual PC follows the same rule starting from
$\{4,8,16,32\}$, giving buckets $\{4,8\}$, $\{4,8,16\}$,
$\{4,8,16,32\}$, and $\{4,8,16,24\}$ for Cube-Single,
Cube-Double, Cube-Triple, and Scene, respectively.

\paragraph{Benchmark reporting.}
\label{app:benchmark-reporting}
The visual manipulation results use the \texttt{*-play-v0} datasets;
the baselines in this group use four training seeds.
HorizonFlow uses the same configuration for Maze2D and Multi2D.
Its uncertainties are standard deviations across training-seed means.
For HD in \ctabref[(a)]{tab:benchmarks}, we convert the reported standard
errors over 100 planning seeds~\citep{chen2024simple}
to standard deviations using $\mathrm{SD}=\mathrm{SE}\sqrt{100}$.
These values are approximate because the published standard errors
are rounded, and reflect planning-seed rather than training-seed variability.
HDMI and SSD uncertainties in panel (a) likewise correspond to reported
standard errors converted using $\mathrm{SD}=\mathrm{SE}\sqrt{5}$
~\citep{li2023hdmi,kim2024ssd}.
VHD results in panel (a) are from our reproduction and report standard
deviations. DF uncertainties in panel (a) are standard deviations, as specified in
the published paper's statistical-significance checklist
~\citep{chen2024diffusionforcing}.

SAW results in panels (b,c) are taken from Table~1 of
\citet{zhou2025saw}, which reports means and standard deviations
over eight training seeds for navigation and four for visual manipulation.

\paragraph{Sources of published results.}
In \ctabref[(a)]{tab:benchmarks}, Diffuser and DF results are taken
from Table~1 of \citet{chen2024diffusionforcing}; HDMI and SSD
results are taken from Table~1 of \citet{kim2024ssd}; and HD results
are taken from Tables~1 and~7 of \citet{chen2024simple}.
VHD is reproduced by us, rather than copied from a published table.
In panels (b,c), QRL, CRL, HIQL, GCIVL, and GCIQL results, wherever
reported, are taken from the full benchmark table (Table~2) of
\citet{park2025ogbench}. SAW uses the published results described
above. CTA results for AntMaze and HumanoidMaze are taken from
Table~6 of \citet{kim2026cta}, and its visual Cube-Single,
Cube-Double, and Scene results from their Table~7.
CTA results for all three PointMaze environments and visual
Cube-Triple are our reproductions. Diffuser, HD, and DF results in
panel (b) are also our reproductions. Thus, HD's planning-seed
uncertainty convention applies only to panel (a); its panel (b)
uncertainties are standard deviations across four training seeds.

\paragraph{Shared HorizonFlow execution settings.}
HorizonFlow's main benchmark evaluations use EMA parameters, 20 Euler steps
per generation, $K_{\mathrm{RP}}=16$, and $K_{\mathrm{PC}}=4$.
Both levels use FK steering after Euler steps 3 and 7 with
$\beta=0.5$, followed by minimum-count selection.
RP is updated every $H_{\mathrm{RP}}$ environment steps, while PC
replans every step and executes the first generated action.
Evaluation uses each environment's episode limit $T$.
Explicit ablations override their swept settings; the diagnostic
protocols described elsewhere are separate from this recipe.

\subsection{Baseline Implementations and Reproduction Protocols}
\label{app:baseline-reproduction}

\paragraph{OGBench planning implementations.}
We adapt the public implementations of Diffuser, HD, and DF
to the OGBench navigation datasets
\citep{janner2022diffuser,chen2024simple,chen2024diffusionforcing}.
Diffuser is ported from the \texttt{maze2d} branch of
\texttt{jannerm/diffuser}, HD from the \texttt{maze\_2d} branch of
\texttt{changchencc/Simple-Hierarchical-Planning-with-Diffusion},
and DF from \texttt{buoyancy99/diffusion-forcing}.
The JAX ports are checked against the reference implementations
using forward, loss, gradient, and sampling comparisons under
matched inputs. These checks concern the ported numerical
operations; the OGBench observation and execution adaptations
are described separately below.

\paragraph{Planning space and execution.}
All three reproduced planners generate plans in the full observation
space. Diffuser generates state sequences with endpoint inpainting.
HD uses endpoint inpainting at both levels, with a high-level stride
of 15 and 16-token low-level segments. Its low-level model predicts
actions and observations, but execution follows the generated
observation waypoints through the shared controller described below.
DF generates observation sequences with reconstruction guidance on
the position coordinates; we retain its original guidance mechanism
rather than use the separately tested goal-inpainting variant.
DF uses frame stacking of 10, pyramid scheduling, 50 DDIM steps,
guidance scale 2, and a replanning interval of 50 environment steps.

\paragraph{Training the planning models.}
Diffuser and HD use temporal U-Nets with base width 64, Adam with
learning rate $2\times10^{-4}$, effective batch size 64, and
two million optimizer updates. They predict clean tokens and use
limits normalization and EMA with decay 0.995. Diffuser uses 256
diffusion steps; HD uses 256 and 128 at its high and low levels,
respectively, with low-level action-loss weight 10.
DF uses a 12-layer transformer of width 128, four attention heads,
feed-forward width 512, and 1,000 diffusion steps. It is trained
for 200,005 updates with batch size 1,024, AdamW learning rate
$5\times10^{-4}$, weight decay $10^{-4}$, and 10,000 warmup steps,
without EMA. Final evaluations use four independently trained
planning models with seeds 43--46.

\paragraph{Choice of low-level controller.}
The original Maze2D waypoint controllers assume position--velocity
observations and planar control, which do not transfer directly to
OGBench's observation and action interfaces. In our controller
comparisons, full-observation plans with a flow-matching controller
tracked more reliably than position-only or inverse-dynamics
alternatives, particularly on HumanoidMaze. We use this same
controller across the reproduced planners to hold their low-level
execution module fixed.

\paragraph{Shared low-level controller.}
The three planning baselines share one separately trained
goal-conditioned flow-matching controller per environment, with
the same controller checkpoint used across planning-model seeds.
It is conditioned on the current observation and a full-observation
waypoint, not on HorizonFlow's learned latent representation.
The controller uses a three-layer MLP with 512 hidden units per
layer and SiLU activations. Training uses one million updates,
batch size 1,024, learning rate $8\times10^{-4}$, and seed 43;
hindsight goal offsets are sampled uniformly from 1 to 32 steps.
Execution uses 20 Euler steps. Reported baseline standard deviations
therefore measure variation across planning-model training seeds
conditional on this shared controller, rather than variation from
independently retraining the controller for every seed.

\paragraph{Horizon search.}
We explore training horizons around one quarter, one half, and the
full environment episode limit using tuning seed 99, with
architecture-compatible lengths for the temporal U-Nets.
The initial Diffuser and HD screening runs use 500,000 updates;
DF screening uses its full training budget. The initial horizon
screen uses evaluation episode indices 40--49 across the five tasks.
The final evaluation later uses indices 0--49, so these initial
screening indices are not disjoint from the final evaluation
indices, although the tuning-model seed is separate; this overlap can
only favor the tuned baselines.
The final selected horizon parameters are listed in
\Cref{tab:baseline-horizons}; for HD the parameter specifies the
high-level temporal span, and its token count is the ceiling of
this span divided by 15. DF's parameter is its training window
length, not its replanning interval. The full-length DF candidate
for HumanoidMaze Giant was excluded because it exceeded GPU memory.

\begin{table}[t]
\centering
\caption{Selected training horizon parameters for reproduced
OGBench planning baselines, in environment steps. HD uses a
high-level span parameter; DF uses a training window length.}
\label{tab:baseline-horizons}
\begin{tabular}{lrrr}
\toprule
Environment & Diffuser & HD & DF \\
\midrule
PointMaze Medium & 256 & 500 & 250 \\
PointMaze Large & 500 & 500 & 500 \\
PointMaze Giant & 500 & 1000 & 500 \\
AntMaze Medium & 500 & 500 & 500 \\
AntMaze Large & 500 & 500 & 1000 \\
AntMaze Giant & 500 & 500 & 250 \\
HumanoidMaze Medium & 1000 & 1000 & 500 \\
HumanoidMaze Large & 1000 & 1000 & 500 \\
HumanoidMaze Giant & 1000 & 2000 & 1000 \\
\bottomrule
\end{tabular}
\end{table}

\paragraph{Inference-setting search and evaluation.}
Subsequent inference sweeps examine replanning intervals,
inference horizon caps, waypoint advancement and lookahead,
EMA use, controller integration steps, and DF guidance scales.
Candidates are screened with seed 99 on 100 episodes, then compared
with the default setting using seeds 43--46 on episode indices
50--74 of each task (500 episodes in total). This confirmation set
is disjoint from the final evaluation indices 0--49.
The recorded selection rule retains changes with a two-proportion
$z$ statistic of at least 2. The selected changes are replanning
every 500 steps for Diffuser on PointMaze Giant and HumanoidMaze
Giant, every 400 steps for Diffuser on AntMaze Large, and every
500 steps for HD on PointMaze Large. Other Diffuser and HD settings
retain one-shot planning; DF retains its 50-step update interval.
Each final planning model is evaluated on 50 episodes per task,
or 250 per environment. We report the mean and sample standard
deviation of the four seed-level success rates. Aggregate means
are calculated before rounding the displayed per-environment values.

\paragraph{CTA implementation and configuration search.}
For the four reproduced CTA settings, we use the authors' public
implementation, \texttt{rllab-snu/CTA}, at commit \texttt{e9f348f0},
with operational changes for logging, checkpoint handling, and
external evaluation rather than a reimplementation of its learning
algorithm. On pilot seed 99, we sweep subgoal intervals
$\{10,25,50\}$ for PointMaze. The selected intervals are 10, 25,
and 50 for Medium, Large, and Giant, respectively.
For visual Cube-Triple, we compare the default interval of 30
with two variants that allow policy gradients into the learned
representation, using intervals 30 and 10. The selected setting
retains the default interval of 30 and stops those gradients.

\paragraph{CTA training and evaluation.}
Selected configurations are trained with eight seeds (43--50)
for PointMaze and four seeds (43--46) for visual Cube-Triple.
State-input runs use one million updates and visual runs use
500,000 updates. Both use batch size 256, Adam learning rate
$3\times10^{-4}$, and transduction latent dimension 8.
The discount is 0.99 except for PointMaze Giant, where it is 0.995.
Visual observations use the IMPALA-small encoder with image
augmentation probability 0.5. We evaluate the final checkpoints
on 50 episodes for each of five tasks, yielding 250 episodes per
environment and training seed. The reported mean and sample
standard deviation are computed across seed-level success rates.

\paragraph{VH-Diffuser implementation and training.}
For Maze2D/Multi2D, we independently implement VH-Diffuser
\citep{liu2025vhdiffuser} in JAX and use the standard D4RL
\texttt{maze2d-*-sparse-v1} datasets with trajectory-boundary-aware
sampling. The diffusion model predicts normalized action--state
trajectories with six channels, endpoint conditioning, and a
clean-sample prediction objective. Its Diffuser-style temporal
U-Net is trained for two million updates with batch size 32,
Adam learning rate $2\times10^{-4}$, gradient clipping at 1,
EMA decay 0.995, and a 256-step cosine diffusion schedule.
Variable-length training crops use 32-step buckets up to maximum
horizons of 128, 256, and 384 for U-Maze, Medium, and Large.

\paragraph{VH-Diffuser length-predictor search.}
Our length-predictor implementation uses 64 random Fourier features
and three 512-unit LayerNorm/ReLU hidden layers with a positive
scalar output. It is trained with the method's TD-style length
learning formulation, Adam learning rate $3\times10^{-4}$,
gradient clipping at 1, and target EMA decay 0.995.
We retain the 20,000-update predictor for U-Maze and train the
Medium and Large predictors for 500,000 updates while keeping
their diffusion models fixed. The selected anchor-step grids are
$\{1,2,4,8,16,32,64,128,256\}$ for Medium and
$\{1,2,4,8,16,32\}$ for U-Maze and Large.
Predictor-training and inference variants were explored initially
on seed 0 and then expanded to five training seeds; this development
procedure did not use a separately recorded held-out tuning split,
which can only favor the reproduced baseline.

\paragraph{VH-Diffuser inference adaptations.}
Predicted horizons are rounded up to a multiple of 32 and clipped
to the range from 32 to the environment-specific maximum.
We additionally use two reproduction-side adaptations, rather than
attribute them to the original algorithm. Horizon search doubles
the horizon and regenerates a plan when the 90th percentile of
consecutive waypoint displacements exceeds 0.06 in unnormalized
position coordinates, subject to the maximum horizon and at most
three expansions. Exhaustion-triggered replanning predicts a new
horizon from the current state if the goal has not been reached;
its backoff imposes a lower bound of twice the exhausted plan's
length, capped by the maximum horizon.
U-Maze uses this replanning rule only for the single-goal protocol.
Medium and Large use horizon search for both protocols and
exhaustion-triggered replanning only for the single-goal protocol.
Each generation call uses a single candidate, without value-based
or execution-based candidate selection. Plans are followed by the
position--velocity waypoint controller, with actions clipped to
$[-1,1]$. Evaluation uses episode limits of 300, 600, and 800 steps
for U-Maze, Medium, and Large, and 100 episodes per training seed
and protocol across five seeds. The table reports normalized-score
means and sample standard deviations across the seed-level means.
The horizon and selection diagnostics use separate protocols
described in \Cref{app:horizon-diag}.

\clearpage
\subsection{Training Time and Hardware}
\label{app:training-time}

\ctabref{tab:training-time} reports wall-clock time for one training
run per seed on each dataset. HorizonFlow timings include both the
route planner and prefix controller, with the two stages reported
separately. Navigation and Maze2D runs use an RTX~5090; visual
manipulation runs use an H200.

\begin{table}[htbp]
\centering
\setlength{\tabcolsep}{4pt}
\renewcommand{\arraystretch}{1.08}
\caption{Wall-clock training time per seed in hours.
HorizonFlow entries report total time (RP + PC).
Dashes indicate unreported measurements.}
\label{tab:training-time}
\begin{tabular*}{\linewidth}{@{\extracolsep{\fill}}lcrrrr@{}}
\toprule
Dataset & GPU & HorizonFlow & Diffuser & HD & DF\\
\midrule
PointMaze-Medium    & RTX 5090 & 12.1 (4.8 + 7.2) & 4.9  & 4.2 & 0.8\\
PointMaze-Large     & RTX 5090 & 12.1 (4.8 + 7.2) & 6.8  & 4.1 & 1.4\\
PointMaze-Giant     & RTX 5090 & 12.2 (4.8 + 7.3) & 6.8  & 4.4 & 1.4\\
AntMaze-Medium      & RTX 5090 & 12.1 (4.8 + 7.2) & 6.8  & 4.1 & 1.5\\
AntMaze-Large       & RTX 5090 & 12.1 (4.8 + 7.2) & 6.8  & 4.1 & 3.6\\
AntMaze-Giant       & RTX 5090 & 12.1 (4.8 + 7.2) & 6.9  & 4.1 & 0.8\\
HumanoidMaze-Medium & RTX 5090 & 12.1 (4.8 + 7.2) & 13.0 & 4.5 & 1.5\\
HumanoidMaze-Large  & RTX 5090 & 12.1 (4.8 + 7.2) & 13.0 & 4.5 & 1.5\\
HumanoidMaze-Giant  & RTX 5090 & 12.1 (4.8 + 7.2) & 13.0 & 5.1 & 3.7\\
\midrule
Maze2D-U-Maze       & RTX 5090 & 8.8 (4.8 + 4.0)  & -- & -- & --\\
Maze2D-Medium       & RTX 5090 & 10.1 (4.8 + 5.3) & -- & -- & --\\
Maze2D-Large        & RTX 5090 & 10.9 (4.8 + 6.0) & -- & -- & --\\
\midrule
Visual Cube-Single & H200 & 11.6 (9.5 + 2.1) & -- & -- & --\\
Visual Cube-Double & H200 & 12.1 (9.5 + 2.6) & -- & -- & --\\
Visual Cube-Triple & H200 & 12.9 (9.4 + 3.4) & -- & -- & --\\
Visual Scene       & H200 & 12.8 (9.5 + 3.3) & -- & -- & --\\
\bottomrule
\end{tabular*}
\end{table}

\paragraph{H200 training hardware.}
Visual-manipulation training uses one NVIDIA H200 with 141\,GB
HBM3e memory (143,771\,MiB reported) and a 700\,W power limit.
The host uses Intel Xeon Platinum~8480+ processors, with
224 logical CPUs visible to the container and approximately
3\,TB RAM (3,023\,GiB reported by the OS).
The software environment is Ubuntu~24.04.4 LTS,
NVIDIA driver~580.159.03 (driver-reported CUDA~13.0),
Python~3.12.3, JAX/jaxlib~0.11.2, Flax~0.12.9,
Optax~0.2.8, and NumPy~2.5.3.

\clearpage
\section{Background on Joint Insertion and Flow Generation}
\label{app:generative-background}

This appendix records the formal background behind the two operations
introduced in \Cref{sec:prelim}. The edit-rate formulation explains
how sequence length can change during generation; the OneFlow details
identify the inherited count decomposition and local-time construction.
The control hierarchy and selection rule are described independently
in \Cref{sec:method}.

\subsection{Edit Flows: Variable-Length Structure}
\label{app:edit-flows}

Edit Flows generalizes discrete flow matching to the sequence space
$\mathcal{X}=\bigcup_{n=0}^{N}\mathcal{V}^{n}$, where $\mathcal{V}$ is a
discrete token vocabulary, $\mathcal{V}^{n}$ is the set of sequences of
length $n$, and $N$ is the maximum allowed length. Thus,
$\mathcal{X}$ contains sequences of every length from zero (the empty
sequence) to $N$. Edit Flows defines a CTMC whose jumps are insertions,
deletions, or substitutions~\citep{havasi2025editflows}.
Restricting the admissible edit set to insertions yields an
insertion-only process. Let $\mathcal{E}(x)$ denote the admissible edits
from $x$, and let $e(x)$ be the
sequence after applying edit $e$. The model
assigns each edit a nonnegative rate
$r_\theta(e\mid x,t)$, with infinitesimal transition probability
\begin{equation}
  \Pr\!\left[
    X_{t+\mathrm{d}t}=e(x)
    \mid X_t=x
  \right]
  =
  r_\theta(e\mid x,t)\,\mathrm{d}t
  + o(\mathrm{d}t).
  \label{eq:ctmc}
\end{equation}
The probability of no edit is
$1-\sum_e r_\theta(e\mid x,t)\mathrm{d}t+o(\mathrm{d}t)$.
Consequently, under an insertion-only process the final sequence length is its initial length plus
the number of insertions realized before $t=1$, rather than a fixed
tensor dimension.

Training uses an auxiliary alignment between a source and target
sequence to construct tractable conditional target rates
$\bar{r}_t(e\mid x_t,x_1)$. Up to terms independent of $\theta$,
the rate-matching Bregman objective is
\begin{equation}
  \mathcal{L}_{\mathrm{rate}}
  =
  \mathbb{E}
  \!\left[
    \sum_{e\in\mathcal{E}(x_t)}
      r_\theta(e\mid x_t,t)
    -
    \sum_{e\in\mathcal{E}(x_t)}
      \bar{r}_t(e\mid x_t,x_1)
      \log r_\theta(e\mid x_t,t)
  \right].
  \label{eq:ef}
\end{equation}
Its population optimum is the marginal rate
$r^*(e\mid x_t,t)
=\mathbb{E}[\bar{r}_t(e\mid x_t,x_1)\mid x_t]$.
Because insertions are defined relative to the current sequence, the
model can grow a sequence at any gap without prescribing its final
length. Our implementation uses padded storage with a variable active
count, as described in \Cref{app:method:objects}.

\subsection{OneFlow: Joint Structure and Content}
\label{app:oneflow}

OneFlow combines insertion for discrete text with FM for continuous
image latents in a shared sequence model~\citep{nguyen2025oneflow}.
For each gap, it factorizes the insertion
rate into a missing-count parameter and a distribution over discrete
token identities. To handle the concentration of zero-count targets, it
separates gap completion from the count on nonempty gaps: a Bernoulli
head predicts whether the gap is complete, and the original Poisson
regression loss is applied only when the target count is nonzero.
Suppressing the token-identity loss and parameter-independent terms,
the resulting count objective for a gap $g$ is
\begin{equation}
  \mathcal{L}_{\mathrm{count}}^{\mathrm{OneFlow}}
  = \mathbb{E}_{g}\!\left[
    \mathrm{BCE}\!\left(\pi(g),\mathbf{1}[k_g=0]\right)
    + \mathbf{1}[k_g>0]\left(
      \lambda_{\mathrm{nz}}(g)-k_g\log\lambda_{\mathrm{nz}}(g)
    \right)
  \right].
  \label{eq:oneflow-count}
\end{equation}
OneFlow also assigns separate FM times to continuous blocks introduced
at different stages. An image inserted partway
through generation is initialized from noise and therefore cannot share
the refinement time of older images. With insertion scheduler $\eta$, OneFlow samples
$u\sim\mathrm{Unif}(0,1)$ and assigns an inserted image the local time
\begin{equation}
  \sigma_{\mathrm{img}}
  =
  \sigma_{\mathrm{text}}-\eta^{-1}(u),
  \qquad
  t_{\mathrm{img}}
  =
  \operatorname{clip}_{[0,1]}(\sigma_{\mathrm{img}}),
  \label{eq:oneflow-time}
\end{equation}
treating $\sigma_{\mathrm{img}}<0$ as not yet inserted. Thus each
continuous image block follows its own FM clock while text insertions
continue.
HorizonFlow retains this local-time construction and zero/nonzero
count decomposition, modeling positive missing counts with a
normalized zero-truncated Poisson distribution
(Equation~\ref{eq:ins-loss-main}). The resulting head mean is used in
both the insertion rule and the count-based steering score.

\clearpage
\section{Additional Ablations and Inference Costs}
\label{app:compute-alloc-domains}

\begin{figure}[ht]
  \centering
  \includegraphics[width=\linewidth]{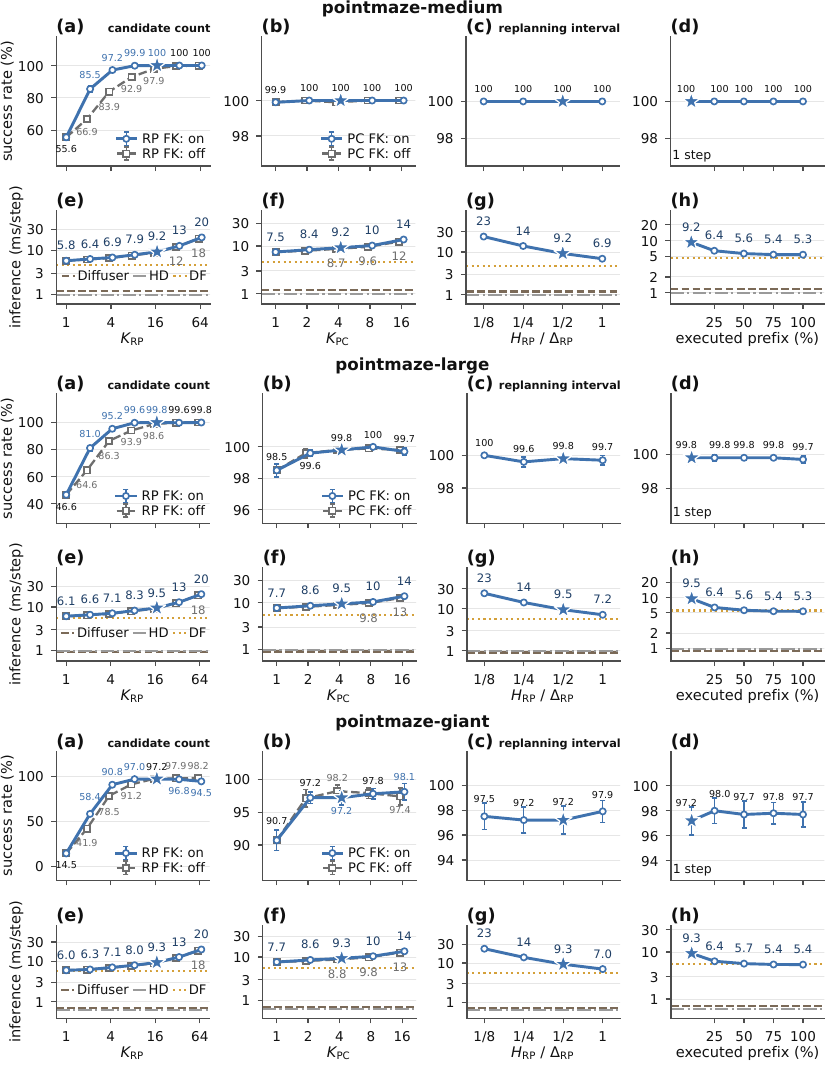}
  \caption{Sampling, replanning, and inference cost on PointMaze.
  Blocks show Medium, Large, and Giant, top to bottom.
  Each block reports success rate (\%) (a--d) and estimated inference time
  (e--h, ms/step, logarithmic scale) for RP and PC candidate counts,
  RP update ratio, and executed prefix fraction, respectively.
  Solid/dashed curves indicate FK steering on/off; stars mark defaults.}
  \label{fig:compute-alloc-pointmaze}
\end{figure}

\begin{figure}[ht]
  \centering
  \includegraphics[width=\linewidth]{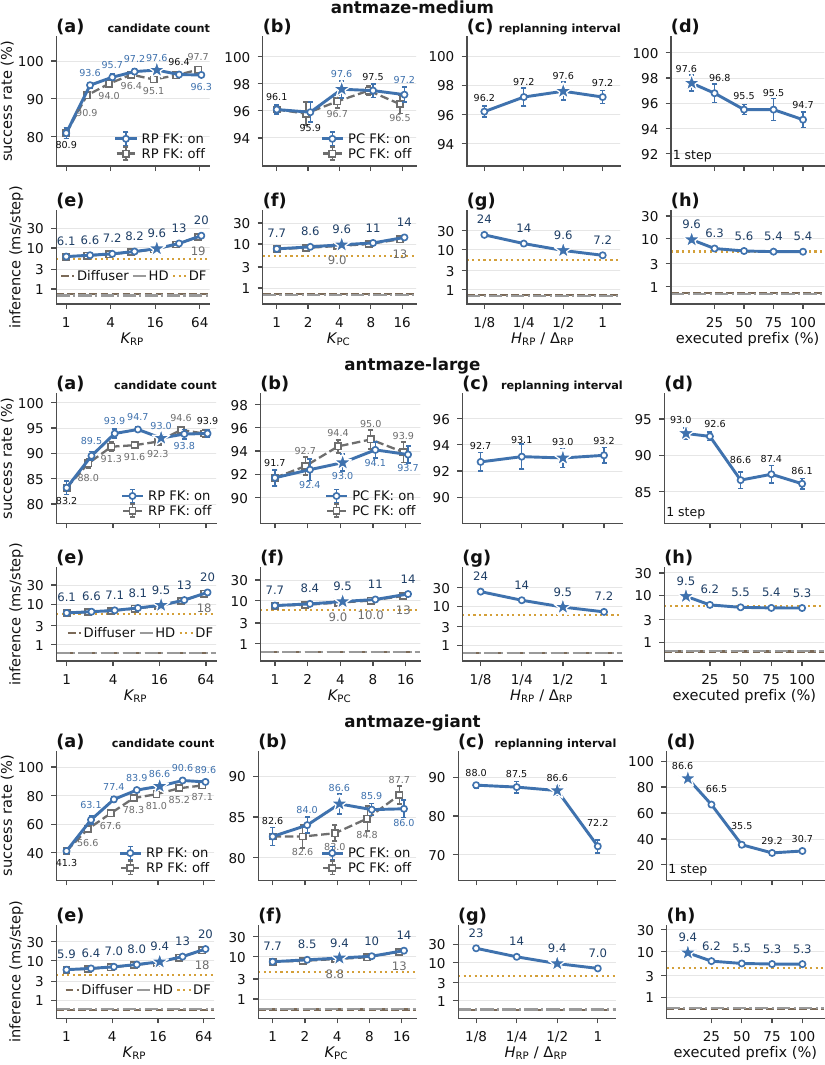}
  \caption{Sampling, replanning, and inference cost on AntMaze.
  Blocks show Medium, Large, and Giant, top to bottom.
  Each block reports success rate (\%) (a--d) and estimated inference time
  (e--h, ms/step, logarithmic scale) for RP and PC candidate counts,
  RP update ratio, and executed prefix fraction, respectively.
  Solid/dashed curves indicate FK steering on/off; stars mark defaults.}
  \label{fig:compute-alloc-antmaze}
\end{figure}

\begin{figure}[ht]
  \centering
  \includegraphics[width=\linewidth]{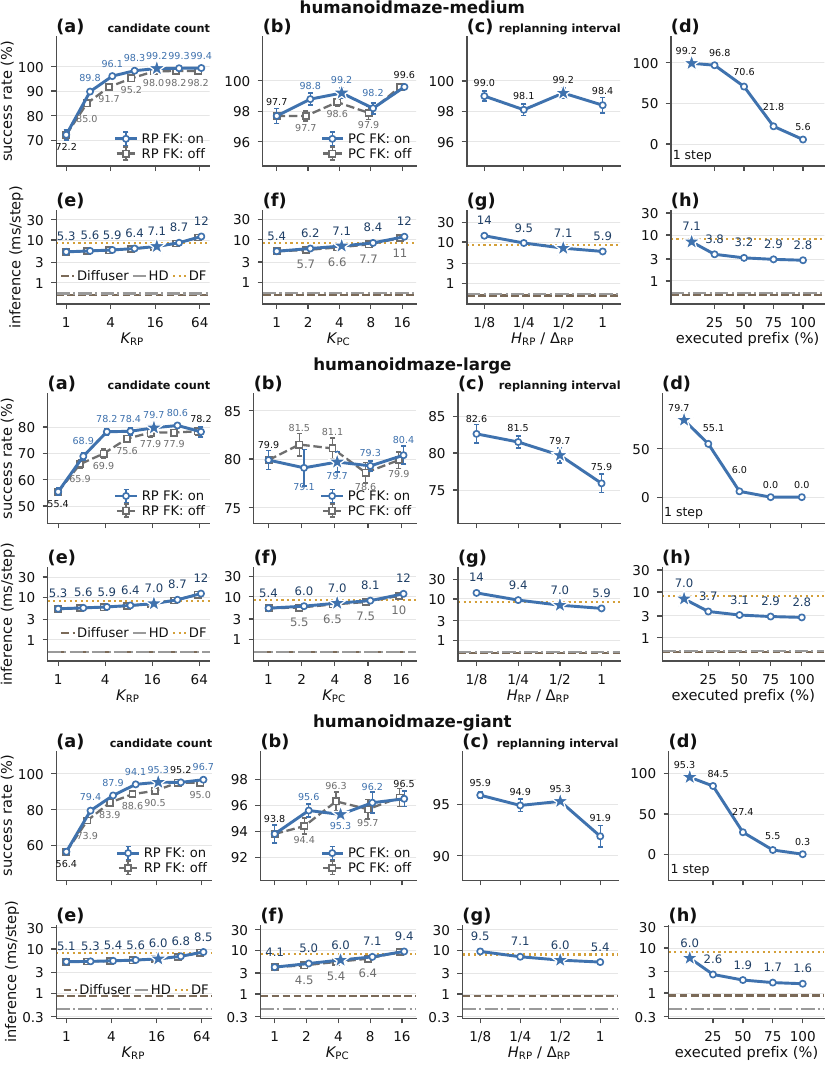}
  \caption{Sampling, replanning, and inference cost on HumanoidMaze.
  Blocks show Medium, Large, and Giant, top to bottom.
  Each block reports success rate (\%) (a--d) and estimated inference time
  (e--h, ms/step, logarithmic scale) for RP and PC candidate counts,
  RP update ratio, and executed prefix fraction, respectively.
  Solid/dashed curves indicate FK steering on/off; stars mark defaults.}
  \label{fig:compute-alloc-humanoidmaze}
\end{figure}

\begin{figure}[ht]
  \centering
  \includegraphics[width=\linewidth]{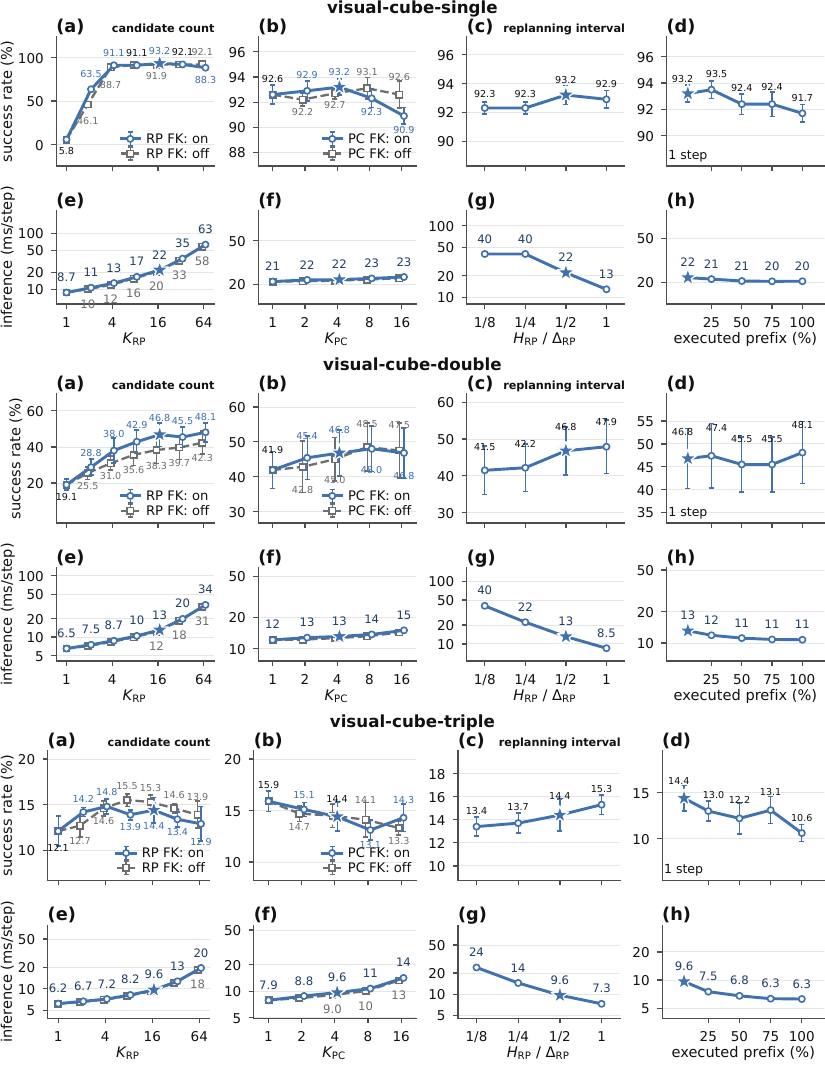}
  \caption{Sampling, replanning, and inference cost on visual cube manipulation.
  Blocks show Cube-Single, Cube-Double, and Cube-Triple, top to bottom.
  Each block reports success rate (\%) (a--d) and estimated inference time
  (e--h, ms/step, logarithmic scale) for RP and PC candidate counts,
  RP update ratio, and executed prefix fraction, respectively.
  Solid/dashed curves indicate FK steering on/off; stars mark defaults.}
  \label{fig:compute-alloc-manipulation}
\end{figure}

\begin{figure}[ht]
  \centering
  \includegraphics[width=\linewidth]{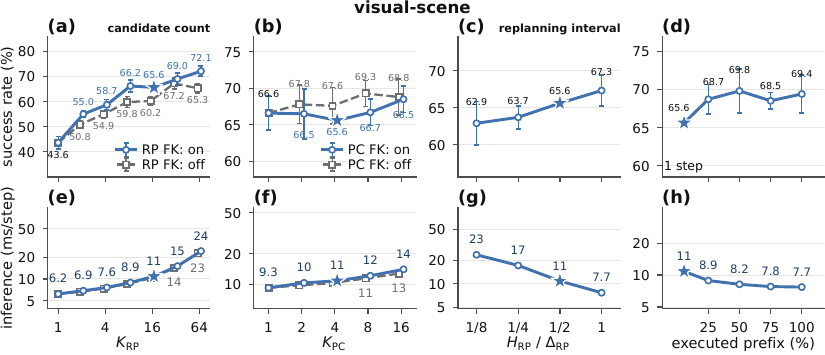}
  \caption{Sampling, replanning, and inference cost on visual Scene.
  Panels (a--d) report success rate (\%), and (e--h) report estimated
  inference time (ms/step, logarithmic scale), for RP and PC candidate
  counts, RP update ratio, and executed prefix fraction, respectively.
  Solid/dashed curves indicate FK steering on/off; stars mark defaults.}
  \label{fig:compute-alloc-scene}
\end{figure}

\begin{figure}[ht]
  \centering
  \includegraphics[width=\linewidth]{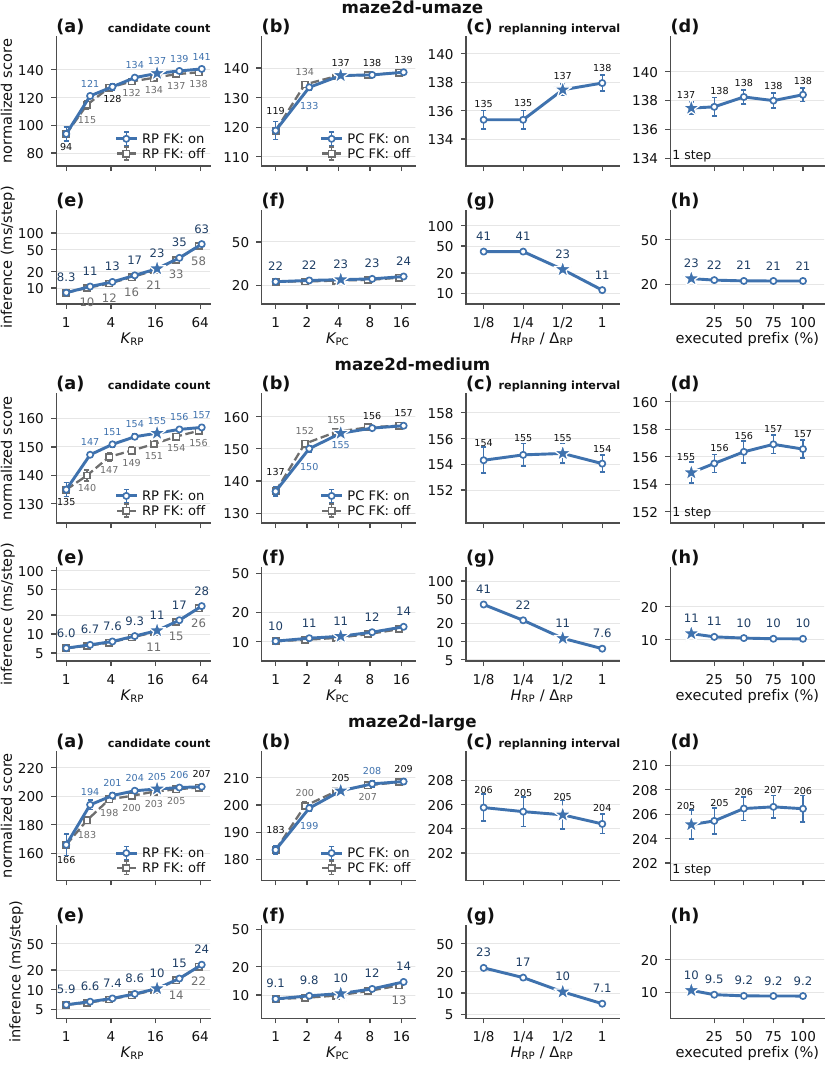}
  \caption{Sampling, replanning, and inference cost on Maze2D.
  Blocks show U-Maze, Medium, and Large, top to bottom.
  Each block reports D4RL normalized score (a--d) and estimated inference time
  (e--h, ms/step, logarithmic scale) for RP and PC candidate counts,
  RP update ratio, and executed prefix fraction, respectively.
  Solid/dashed curves indicate FK steering on/off; stars mark defaults.}
  \label{fig:compute-alloc-maze2d}
\end{figure}

This appendix gives environment-wise sampling, replanning, and
inference-cost results for \cfigref{fig:compute-alloc}, followed by
architecture ablations. The corresponding panels and legends are
identified in \cfigref{fig:compute-alloc-pointmaze}--\ref{fig:compute-alloc-maze2d}.

\paragraph{Evaluation counts and aggregation.}
Navigation sweeps use eight training seeds and 125 evaluation problems
per environment and seed, comprising 25 problems for each of five
tasks. All settings, including the default points, use this matched
subset rather than the larger main-benchmark evaluation set.
Visual manipulation sweeps use four training seeds and 250 problems
per environment and seed, with 50 problems per task. Maze2D sweeps use
five training seeds and 100 single-goal episodes per environment and
seed, reporting D4RL-normalized scores.

Each environment-wise curve reports the mean across training seeds
with error bars of one standard error. For the aggregate
success-rate curves in \cfigref{fig:compute-alloc}, we first average
within each environment across seeds, then give equal weight to all
nine OGBench navigation environments.
Visual manipulation and Maze2D are excluded from this average. If $s_e$ is the
sample standard deviation across seeds for environment $e$, the
aggregate error-bar magnitude is $\sqrt{\sum_{e=1}^{9}s_e^2}/9$.
This propagates the seed standard deviations; it is not a standard
error of the aggregate mean.

\paragraph{Inference-time hardware.}
\label{app:inference-hardware}
We estimate amortized model-inference time from batch-one, device-only
median planner and controller call times, excluding compilation,
environment simulation, and host--device transfers. Call times are
weighted by the corresponding evaluation call counts and divided by
the number of executed environment steps for each environment and
setting. The main-figure aggregate then equally averages the nine
navigation environments. These are average model-compute
estimates, not end-to-end or worst-case control latencies.
For visual PC-execution sweeps, call counts come from a separate
timing probe with five episodes per task and training seed;
the success-rate curves use the full evaluation described above.
The baseline timing configurations use one-shot planning for HD and
for Diffuser except on HumanoidMaze-Giant, where Diffuser replans every
500 steps. DF replans every 25 steps on HumanoidMaze-Medium/Large
and every 50 steps elsewhere. These timing configurations
differ from the benchmark configurations in \Cref{app:baseline-reproduction}
for Diffuser on PointMaze-Giant and AntMaze-Large, and HD on PointMaze-Large;
these timing configurations replan less often, which lowers the baselines'
measured cost.

All reported inference-time measurements, including the
visual-manipulation and Maze2D panels, use an NVIDIA GeForce RTX~5090
with 32\,GB GDDR7 memory (32,607\,MiB reported) at the default
600\,W power limit. Reported maximum SM and memory clocks are
3,105\,MHz and 14,001\,MHz, respectively.
The host has one AMD Ryzen~9~9950X CPU (16 cores, 32 threads) and
128\,GB RAM (122\,GiB reported by the OS).
The software environment is Ubuntu~24.04.4 LTS with Linux~6.8.0,
NVIDIA driver~595.71.05 (driver-reported CUDA~13.2),
Python~3.12.3, JAX/jaxlib~0.11.2, and Flax~0.12.9.

\paragraph{Sweep conventions.}
The RP candidate counts are $\{1,2,4,8,16,32,64\}$ and the PC counts
are $\{1,2,4,8,16\}$. The PC sweep holds $K_{\mathrm{RP}}=16$,
and the RP sweep holds $K_{\mathrm{PC}}=4$.
The one-candidate point is shared between the two steering curves.
The RP update ratios are $\{1/8,1/4,1/2,1\}$.
These ratios are nominal: the stride multiplied by the ratio is
rounded to the nearest integer and clamped to at least one step. For Maze2D
U-Maze/Medium/Large, the RP strides are 5/10/13 and the default update
periods are 2/5/6 steps. For U-Maze, the $1/8$ and $1/4$ settings
both map to a one-step period and reuse that evaluation.
Visual cube-single/cube-double/cube-triple/scene
use strides 4/8/16/12 and default periods 2/4/8/6.

Fractional PC execution uses $\max(1,\lceil f n\rceil)$ requested
actions for fraction $f$ and generated non-anchor count $n$, subject
to the available prefix and subsequent RP updates. The fractions
are 25\%, 50\%, 75\%, and 100\%. The separately labeled one-step
point executes one action before replanning; its horizontal plotting
position does not denote a fixed fraction of every generated prefix.

\subsection{Architecture Ablations}
\label{app:architecture-ablation}

\begin{figure}[htbp]
  \centering
  \includegraphics[width=0.46\linewidth]{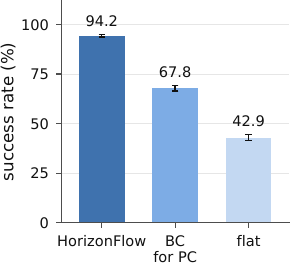}
  \caption{Architecture ablations on OGBench navigation.
  HorizonFlow is compared with replacing the prefix controller by
  a behavior-cloning policy (BC for PC) and removing the hierarchy (flat).
  Bars report mean success across navigation environments and training
  seeds; error bars show standard deviations across seed-level averages.}
  \label{fig:hier-flat}
\end{figure}

\paragraph{Why both a route planner and a prefix controller?}
\cfigref{fig:hier-flat} compares HorizonFlow with two architectural
ablations. Replacing the prefix controller with a behavior-cloning
(BC) policy while retaining the route planner substantially reduces
average success. Removing the hierarchy yields a larger reduction:
the flat planner generates actions directly toward the final goal,
without an intermediate subgoal route. These comparisons support the combined RP--PC design in the evaluated navigation environments.

\paragraph{Configurations and evaluation.}
\cfigref{fig:hier-flat} compares three configurations on the nine
OGBench navigation environments. The BC replacement retains the RP
and replaces PC with a conditional flow-matching MLP that generates
one action from the current state and RP subgoal latent. It is trained
by behavior cloning without advantage weighting, using hindsight
goal offsets sampled uniformly from one to twice the RP training
stride. The flat planner removes the intermediate RP and generates
actions conditioned directly on the final goal, replanning every
step. HorizonFlow uses $K_{\mathrm{RP}}=16$ and $K_{\mathrm{PC}}=4$;
the BC replacement retains the same RP candidate count.
The flat configuration uses 16 candidates.
All three configurations use steering checkpoints 3 and 7
with $\beta=0.5$ for their generative planning components.

HorizonFlow and BC each use eight training seeds; the flat planner
uses three. All three configurations are evaluated on 250 episodes
per environment and training seed.
For each seed, we first average the success rates equally across
all nine environments. Bars report the mean of these seed-level
averages, and error bars report their sample standard deviations.

\clearpage
\subsection{Step-Level Inference Latency}
\label{app:step-latency}

\ctabref{tab:step-latency} distinguishes controller-only steps from
steps that also invoke the planner on OGBench navigation.
Measurements use the batch-one, device-only setup in
\Cref{app:inference-hardware}. After three warm-up calls, we time
50 calls per planning unit and HorizonFlow controller, and 200 calls
per baseline tracking controller, synchronizing device completion.
The lower endpoint is the smallest timed controller-only call.
The upper endpoint adds the largest timed planner and controller
calls, measured separately; for DF, we take the largest planner time
across the tested remaining-horizon settings. HD's planning unit
already includes its two planning levels.
These are component-based estimates, not extrema of jointly measured
environment-step latencies or guaranteed worst-case bounds.

\begin{table}[htbp]
\centering
\small
\setlength{\tabcolsep}{3pt}
\renewcommand{\arraystretch}{1.1}
\caption{Step-level device-inference latency estimates on OGBench
navigation, in milliseconds. Each entry gives the controller-only
minimum and the constructed planning-update upper estimate.
The reference control period is included for comparison.}
\label{tab:step-latency}
\begin{tabular*}{\linewidth}{@{\extracolsep{\fill}}lcrrrr@{}}
\toprule
Environment & \shortstack{Control\\period} & Diffuser & HD & DF & HorizonFlow\\
\midrule
PointMaze-Medium & 100 & 0.27--236.57 & 0.26--285.88 & 0.26--217.11 & 4.54--41.54\\
PointMaze-Large  & 100 & 0.27--248.96 & 0.27--285.73 & 0.28--293.53 & 4.76--42.13\\
PointMaze-Giant  & 100 & 0.27--249.04 & 0.27--297.68 & 0.27--296.08 & 4.65--42.14\\
\midrule
AntMaze-Medium & 100 & 0.27--250.56 & 0.27--289.47 & 0.28--290.66 & 4.81--42.75\\
AntMaze-Large  & 100 & 0.26--250.26 & 0.28--287.69 & 0.28--422.73 & 4.76--42.71\\
AntMaze-Giant  & 100 & 0.28--249.87 & 0.27--289.06 & 0.28--217.24 & 4.61--42.58\\
\midrule
HumanoidMaze-Medium & 25 & 0.31--265.95 & 0.31--304.24 & 0.31--254.75 & 4.68--43.21\\
HumanoidMaze-Large  & 25 & 0.31--266.14 & 0.32--303.75 & 0.31--217.53 & 4.66--43.19\\
HumanoidMaze-Giant  & 25 & 0.32--265.95 & 0.32--358.25 & 0.32--445.83 & 4.79--42.58\\
\bottomrule
\end{tabular*}
\end{table}

HorizonFlow's step-level timing uses the default RP update period,
one-step PC execution, and the candidate counts and FK settings in
\ctabref{tab:recipe}.

The step-level comparison shows why amortized cost alone does not
characterize the computation required at a planning update.
The baseline tracking controllers are inexpensive between updates,
but their planning calls dominate the upper estimates. HorizonFlow
spends more on local action generation at each step while requiring
less computation when updating the route. For one-shot configurations,
the planning cost occurs at initialization rather than throughout
execution; the update frequency therefore matters alongside its cost.

HorizonFlow's planning-update estimates fall below the PointMaze and
AntMaze reference periods. On HumanoidMaze, all planners' planning-update
estimates exceed the 25\,ms reference period, with HorizonFlow's (about
43\,ms) the lowest; controller-only estimates remain below the reference
period in all three domains.
Compilation, environment simulation, and host--device transfers
are excluded, so end-to-end deadline compliance requires separate
measurement.

\clearpage

\clearpage

\clearpage

\clearpage

\clearpage

\clearpage

\end{document}